\documentclass[10pt,twocolumn,letterpaper]{article}
\usepackage[pagenumbers]{cvpr}
\usepackage{times}

\usepackage{comment}

\usepackage{graphicx}
\usepackage{subcaption}
\usepackage{float}

\usepackage{booktabs}                   
\usepackage{multirow}
\usepackage{array}
\usepackage{paralist}
\usepackage{enumitem}

\usepackage{mathtools}
\usepackage{amssymb,amsmath}   

\usepackage{pifont}
\usepackage{xspace}

\usepackage{diagbox}
\usepackage{xcolor}
\usepackage[export]{adjustbox}
\usepackage{sidecap}

\usepackage{multirow, makecell, booktabs}
\usepackage[pagebackref=true,breaklinks=True,colorlinks=true,bookmarks=false,citecolor=blue,linkcolor=blue]{hyperref}

\def\eg{e.g.,~}               
\def\ie{i.e.,~}               

\newcommand{\Paragraph}[1]
{\vspace{1mm} \noindent\textbf{#1}}

\newcommand{\figref}[1]{Figure~\ref{fig:#1}} 
\newcommand{\tabref}[1]{Table~\ref{tab:#1}}

\long\def\ignorethis#1{}

\newlength\paramargin
\newlength\figmargin
\newlength\subfigmargin
\newlength\secmargin
\newlength\subsecmargin
\newlength\tabmargin
\newlength\eqmargin

\newcolumntype{C}[1]{>{\centering\let\newline\\\arraybackslash\hspace{0pt}}m{#1}}

\newcommand*\colourmark[1]{%
  \expandafter\newcommand\csname #1xmark\endcsname{\textcolor{#1}{\ding{56}}}%
}

\colourmark{blue}
\colourmark{red}

\newcommand*\colourchecksnow[1]{%
  \expandafter\newcommand\csname #1snow\endcsname{\textcolor{#1}{\ding{100}}}%
}

\colourchecksnow{cyan}

\definecolor{darkspringgreen}{rgb}{0, 0.6, 0.3}
\newcommand*\colourcheck[1]{%
  \expandafter\newcommand\csname #1check\endcsname{\textcolor{#1}{\ding{52}}}%
}

\colourcheck{blue}
\colourcheck{cadmiumgreen}

\newcommand*\colourtri[1]{%
  \expandafter\newcommand\csname #1tri\endcsname{\textcolor{#1}{\ding{115}}}%
}

\colourtri{black}

\makeatletter
\@namedef{ver@everyshi.sty}{}
\makeatother
\usepackage{tikzsymbols}
\newcommand*\colourcheckfire[1]{%
  \expandafter\newcommand\csname #1fire\endcsname{\textcolor{#1}{\Fire}}%
}
\colourcheckfire{red}

\definecolor{cadmiumgreen}{rgb}{0.0, 0.42, 0.24}

\newcommand{\lavp}{\textsc{LAVisH}}
\newcommand{\lav}{\textsc{LAVisH }}

\begin{document}


\title{Vision Transformers are Parameter-Efficient Audio-Visual Learners}

\author{
Yan-Bo Lin\quad\quad
Yi-Lin Sung\quad\quad
Jie Lei\quad\quad
Mohit Bansal \quad\quad
Gedas Bertasius
\\
Department of Computer Science, UNC Chapel Hill \\
\texttt{\{yblin,ylsung,jielei,mbansal,gedas\}@cs.unc.edu} \\
}

\maketitle
\begin{abstract}

Vision transformers (ViTs) have achieved impressive results on various computer vision tasks in the last several years.
In this work, we study the capability of frozen ViTs, pretrained only on visual data, to generalize to audio-visual data without finetuning any of its original parameters. 
To do so, we propose a latent audio-visual hybrid (\lavp) adapter that adapts pretrained ViTs to audio-visual tasks by injecting a small number of trainable parameters into every layer of a frozen ViT. 
To efficiently fuse visual and audio cues, our \lav adapter uses a small set of latent tokens, which form an attention bottleneck, thus, eliminating the quadratic cost of standard cross-attention. 
Compared to the existing modality-specific audio-visual methods, our approach achieves competitive or even better performance on various audio-visual tasks while using fewer tunable parameters and without relying on costly audio pretraining or external audio encoders.
Our code is available at \url{https://genjib.github.io/project_page/LAVISH/}

\end{abstract}  
\vspace{-5mm}
\section{Introduction}\label{sec:intro}
\vspace{\secmargin}
\begin{figure}[t!]
    \centering
	\includegraphics[width=0.9\linewidth]{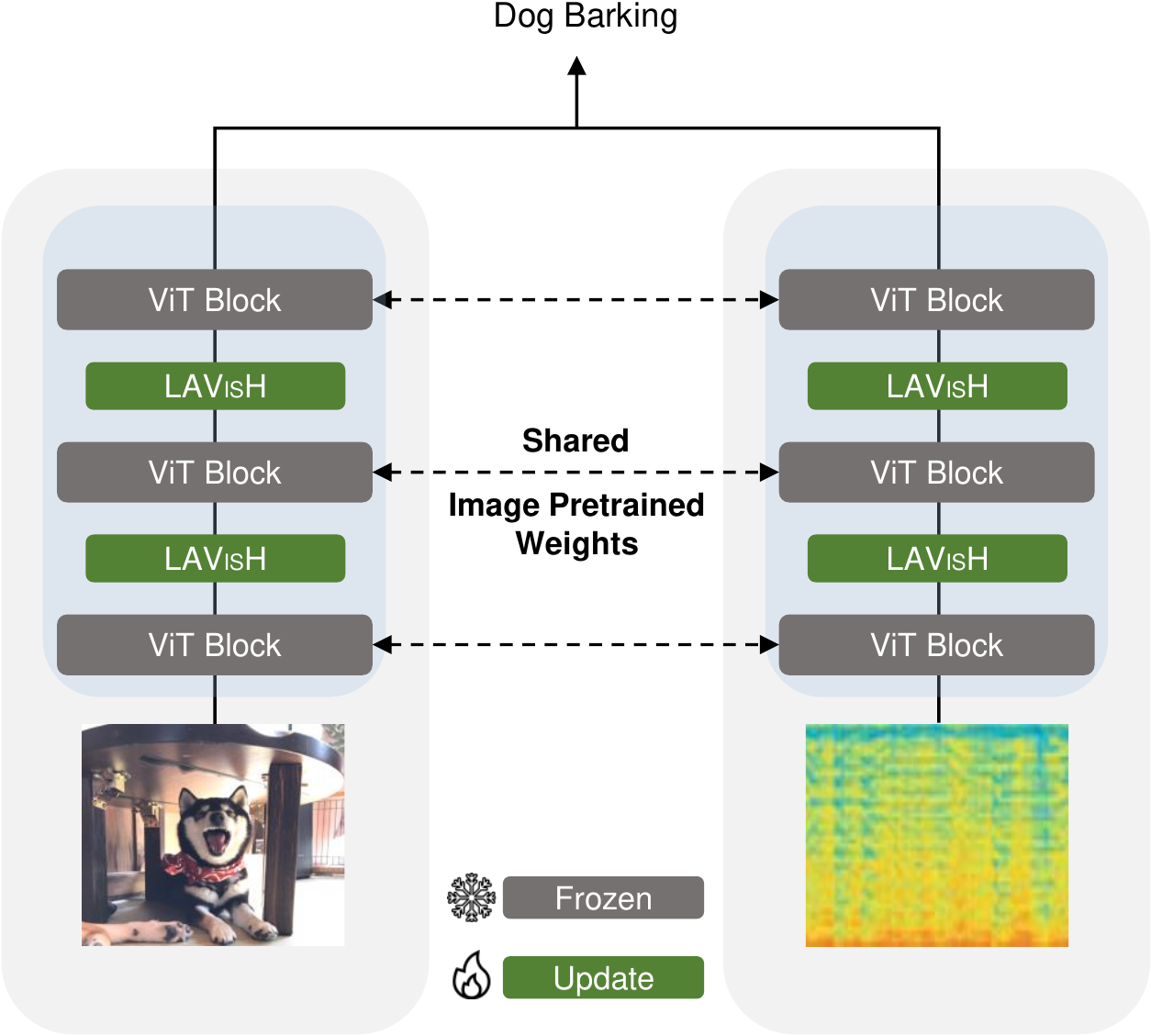}
	\vspace{-2mm}
    \caption{
    We investigate whether frozen vision transformers (ViTs) pretrained only on visual data can generalize to audio data for complex audio-visual understanding tasks.
    For this purpose, we introduce a latent audio-visual hybrid adapter (\lavp), which is inserted into every layer of a frozen ViT model. By tuning only a small number of  additional parameters we can enable a pretrained ViT to efficiently (i) adapt to the audio data, and (ii) fuse relevant cues across audio and visual modalities.
    }
	\label{fig:teaser}
	\vspace{-4mm}
\vspace{\figmargin}
\end{figure}

Humans can seamlessly process audio-visual cues and use them in unison to learn associations between auditory and visual signals (e.g., the sound of \emph{barking} and the visual concept of \emph{dog}). In contrast, most modern computational audio-visual models~\cite{av_iclr21_lee2021crossattentional,av_eccv20_avvp,eccv18_avel,av_cvpr21_av_parsing,cvpr22_ave_cmbs,cvpr22_avqa_avqa,cvpr22_ready_audio_adaptive} study each of these modalities in isolation, which leads to individually-tailored modality-specific models.  While such modality-specific approaches often achieve state-of-the-art results on various audio-visual benchmarks, they also have several major shortcomings. First, optimizing and training models for a specific modality (e.g., audio or video) requires significant research effort and computing power.  %
For example, training large-scale models for audio and video requires more than 2,000 and 5,000 V100 hours respectively~\cite{nips22_audioMAE, icml21_timesformer}, which is not feasible for many smaller research labs. Additionally, since modern visual and audio models are becoming larger, it can be quite costly to use separate backbone networks for processing each modality. For instance, the audio-visual MBT-Large model~\cite{nips21_bottleneck}, built using separate audio and visual encoders, requires more than $48$ GB of GPU memory, which is only available on the costly, high-end GPU servers such as A100. Lastly, the modality-specific approaches are only trained on individual modalities and then typically combined via late fusion. As a result, such models cannot benefit from cross-modal cues in the early layers, which often leads to suboptimal performance on audio-visual tasks requiring joint audio-visual reasoning.

The recent emergence of transformer models~\cite{cvpr22_omnivore,arxiv21_polyvit,arxiv22_uavm,nips21_vatt,nips21_bottleneck} has propelled research in modality-agnostic architectures for multi-modal understanding. In particular, the generality of the transformer architecture~\cite{iclr21_vit} makes it easy to apply these models to different modalities without any modality-specific adaptations. This property is well illustrated by the fact that transformers~\cite{iclr21_vit} currently define state-of-the-art across many domains, including natural language processing (NLP)~\cite{icml21_clip,cvpr22_swinbert,arxiv_clip4clip,iccv21_Frozen,BERT,eccv22_tip_adapter,eccv22_eclipse,eccv22_nagrani2022learning}, computer vision (CV)~\cite{arxiv22_omnimae,icml21_timesformer,iccv21_vivit}, audio analysis~\cite{interspeech21_AST,TASLP_AST,nips22_audioMAE}, speech processing~\cite{interspeech22_mae_ast,nips22_tvlt_textless,iclr22_avhubert}. Such an architecture convergence across different domains/modalities inspired several recent works to investigate the cross-modal generalization of pretrained transformers~\cite{arxiv21_universal_engines,cvpr22_vl_adapter,nips22_ST_Adapter,arxiv21_polyvit}. However, most of them are either focused on language models~\cite{cvpr22_vl_adapter,arxiv21_universal_engines,nips22_few_pe}, or study close-domain transfer (\eg image $\rightarrow$ video)~\cite{arxiv22_omnimae,cvpr22_omnivore,nips22_ST_Adapter}.

In this work, we focus on the cross-modal generalization of pretrained vision transformers (ViT)~\cite{iclr21_vit} to the audio-visual data. Our main inspiration for this study stems from the fact that audio can be represented as a 2D spectrogram, which summarizes 1D raw audio signal into a 2D structure akin to audio images.
Prior work has shown that vision architectures (e.g., CNNs~\cite{icassp17_vggish,icassp20_vggsound} or ViTs~\cite{TASLP_AST,nips22_tvlt_textless}) can be used to process such audio images.
However, most prior methods use these architectures for large-scale audio representation learning.
Instead of pretraining ViTs on large-scale audio data, we hypothesize that the ViTs pretrained on images can simultaneously encode representations that are useful for both images and audio, making them useful for audio-visual tasks without large-scale audio pretraining.

To investigate this hypothesis, we propose a latent audio-visual hybrid (\lavp) adapter that directly adapts frozen ViTs, pretrained only on images, to audio-visual tasks by adding a small number of trainable parameters for audio specialization and audio-visual fusion. Such a scheme allows us to apply frozen ViTs to audio-visual data without updating the original ViT parameters but only the parameters of our proposed \lav modules, which we  insert into every layer of a frozen ViT. For an efficient cross-modal fusion within the \lav module, we use a small set of latent tokens to first compress the information from all modality-specific tokens (e.g., either audio or video) and then apply cross-attention between the latent tokens and all the tokens of another modality (e.g., either video or audio). Such a scheme allows us to eliminate the quadratic cost of standard cross-attention. Furthermore, to allow information transfer between audio-to-video and, conversely, video-to-audio, we adopt a bi-directional \lav scheme, which enables learning a better audio-visual representation.

In our experimental section, we demonstrate that by keeping all the original ViT parameters frozen and updating only a small set of newly added parameters, the frozen ViTs, pretrained only on image data, learn to solve complex audio-visual understanding tasks requiring a joint understanding of audio and visual contents.
In particular, compared to the state-of-the-art modality-specific audio-visual approaches, our method achieves competitive or even better results on the tasks of audio-visual event localization, audio-visual segmentation, and audio-visual question answering while using a smaller number of tunable parameters, and without relying on a separate pre-trained audio encoder (e.g., VGGish~\cite{icassp17_vggish}, AST~\cite{TASLP_AST}, etc.), or costly large-scale audio pretraining. We also show that our proposed latent audio-visual hybrid adapter (\lavp) is more effective and efficient than the standard adapter schemes~\cite{Adapter}.

%

%
%
%

%

\vspace{-2mm}
\section{Related Work}
\vspace{\secmargin}

\Paragraph{Audio-Visual Understanding.} 
Audio-visual understanding tasks focus on the audio-visual perception of objects/events/activities~\cite{av_iclr21_activeContrastive,nips21_av_contrastive,cvpr21_visualvoice,iccv19_cosep,cvpr21_cyclic,av_eccv18_Owens,av_nips18_coop,av_slowfast,nips20_avcluster} using both visual and audio modalities.
For instance, audio-visual event localization~\cite{eccv18_avel,wacv20_ave_avrb,icassp20_ave_avin,iccv19_ave_DAM,aaai20_ave_cman,eccv22_ave_DPNet,wacv23_AVE_CLIP} and audio-visual video parsing~\cite{av_cvpr21_av_parsing,av_eccv20_avvp,my_nips21,eccv22_joint_avvp,nips22_group_avvp} require models for recognizing and localizing joint audio-visual events (e.g., a dog barking).
Most existing approaches~\cite{cvpr22_ave_cmbs,aaai20_ave_cman,cvpr21_ave_psp,eccv22_ave_DPNet} designed for these tasks leverage pretrained modality-specific audio and visual models to extract features and combine them via ad-hoc audio-visual fusion modules.
Moreover, the tasks of sound localization~\cite{av_eccv18_obj_that_sound,av_cvpr18_lls,av_tpami20_lls} and audio-visual segmentation~\cite{eccv22_loc_avs} focus on predicting the regions in the visual scenes corresponding to a sound either using bounding boxes~\cite{av_eccv20_objs_vids,av_nips20_loc,eccv22_loc_ezvsl,nips22_slavc,cvpr22_loc_mix,cvpr21_loc_lvs} or pixel-wise segmentations~\cite{eccv22_loc_avs}.
Most prior sound localization methods tackle this task using self-supervised~\cite{av_cvpr18_lls,eccv22_loc_ezvsl,nips22_slavc,cvpr22_loc_mix} or weakly supervised~\cite{eccv20_loc_MMSL} approaches by learning correspondence between audio and visual patches.
Instead, audio-visual segmentation methods~\cite{eccv22_loc_avs} rely on ground truth masks due to the requirement for precise segmentations. 
Furthermore, the newly introduced audio-visual question answering (AVQA)~\cite{cvpr22_avqa_avqa,cvpr19_avqa_avsd,iccv21_avqa_pano_avqa} task requires methods that perceive both audio and visual modalities to answer human-generated questions about the audio-visual content.
Most methods designed for this task rely on modality-specific audio and vision and models, which are combined via spatial and temporal grounding modules~\cite{cvpr22_avqa_avqa}. 
Unlike these prior methods, which either require modality-specific audio/visual models or expensive pretraining, we study the capability of frozen ViTs, pretrained only on images, to generalize to audio-visual data without any prior large-scale audio-visual pretraining.

\begin{figure*}[t!]
    \centering
	\includegraphics[width=0.9\linewidth]{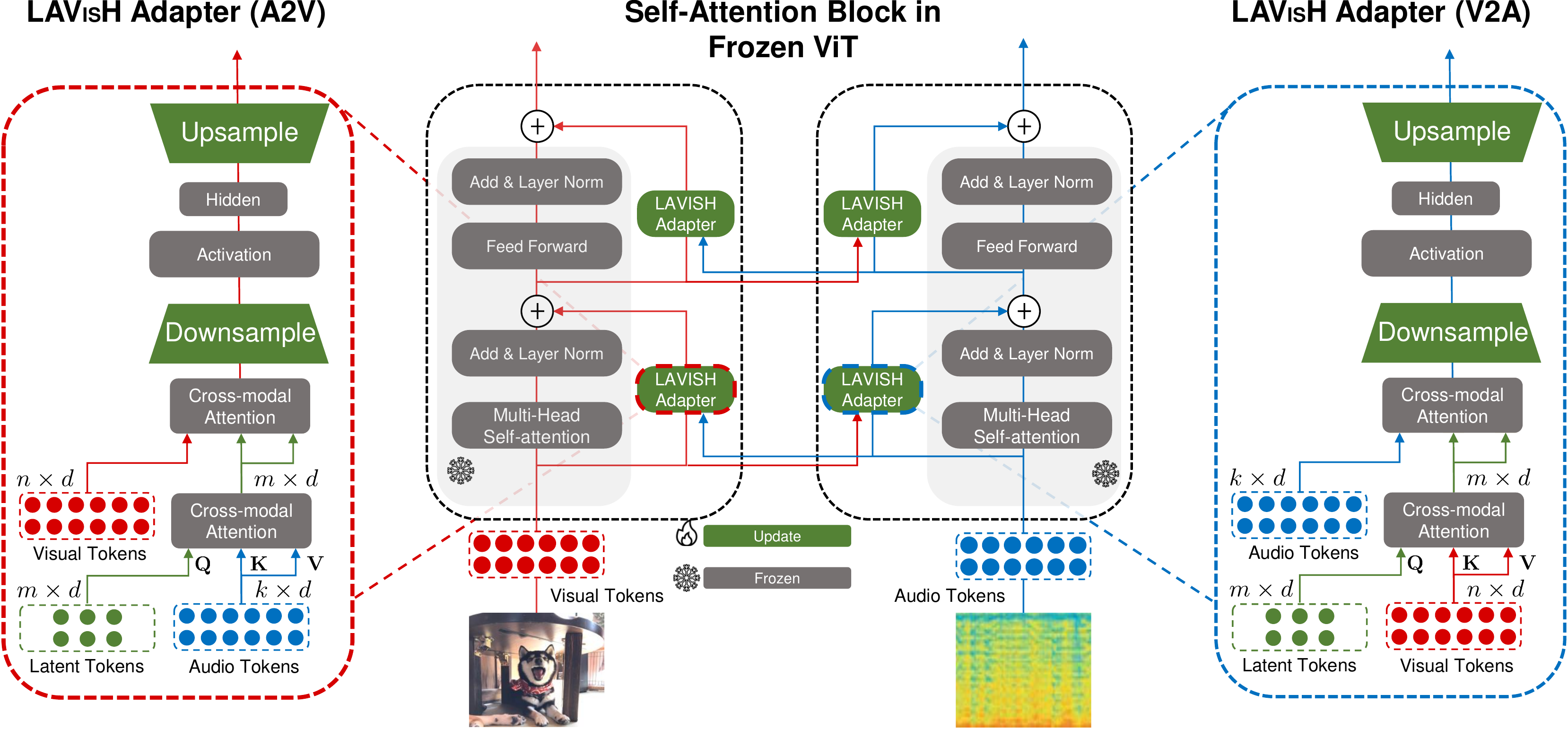}
     \caption{\textbf{Method Overview. Middle:} Our framework consists of a frozen vision transformer (ViT) augmented with trainable latent audio-visual hybrid (\lavp) adapters inserted into each transformer layer.  We use a bi-directional \lav adapter that allows us to transfer information from audio to visual tokens, and conversely from visual to audio tokens.
     \textbf{Left/Right:} Each \lav adapter consists of four high-level components.
     First, we introduce a small number of latent tokens for learning compressed audio or visual representation.   
    Next, the first cross-modal attention operation within the \lav module compresses all the tokens from one modality (either audio or visual) into the latent tokens. 
    Afterward, the second cross-modal attention operation performs audio-visual fusion between the latent tokens of one modality (either audio or visual) and the tokens from another modality (visual or audio).
    Finally, the fused tokens are fed into a lightweight adapter module which computes a more discriminative audio-visual representation and outputs it to the next operation in a ViT layer.\vspace{-0.3cm}
    }
    \vspace{\figmargin}
	\label{fig:method}
\end{figure*}

\Paragraph{Parameter-Efficient Transfer Learning.} 
Parameter-efficient transfer learning aims to adapt pretrained models to new tasks using few trainable parameters.
Most parameter-efficient approaches can be divided into several categories: methods that introduce a small number of additional parameters~\cite{acl21_hyperformer,emnlp21_power,acl21_prefix_tuning}, methods that update only a sparse set of weights in the model~\cite{acl21_parameter_diff_pruning,nips21_training_sparse_masks,acl22_bitfit}, and methods that learn a low-rank factorization of the model's weights~\cite{Lora}.
Adapter~\cite{Adapter} is arguably the most popular parameter-efficient technique among these methods. It consists of lightweight learnable modules inserted between every pair of layers in a pretrained model. Despite their simplicity, adapters achieved impressive results on diverse tasks in both CV~\cite{nips22_polyhistor,eccv22_frozen_clip,arxiv22_vit_adapter,nips17_residual_adapters,cvpr18_efficient_multi-domain,nips22_ST_Adapter} and NLP~\cite{nips22_lst,Adapter,Compacter,alayrac2022flamingo}.
However, most adapter-based approaches are designed for unimodal settings (\ie CV, NLP, etc.), which limits their applications to multi-modal settings since they cannot share cross-modal information.
Recently, several parameter-efficient approaches have been applied to multi-modal settings~\cite{iclr21_parameter_av,cvpr22_vl_adapter}. However, these methods require costly large-scale multimodal pre-training.
Instead, we propose a latent audio-visual hybrid (\lavp) adapter that allows us to adapt frozen ViTs, pretrained only on images, to audio-visual tasks. %

\vspace{-2mm}
\section{Technical Approach}
\vspace{\secmargin}
\label{sec:method}

In this section, we present our proposed latent audio-visual hybrid (\lavp) adapter that  adapts frozen ViTs to audio-visual tasks by updating a small number of additional parameters.
Our proposed \lav module, which we inject into every layer of a frozen ViT, allows the model (i) to adapt to the audio inputs and (ii) fuse information between visual and audio inputs early in the feature representation.
An illustration of our method is presented in~\figref{method}.
Below, we present our technical approach in more detail.
%

\subsection{Audio-Visual Input Embeddings}
\vspace{\subsecmargin}
\label{sec:embedding}
\textbf{Audio and Image Inputs.} Our framework takes audio and visual inputs.
For visual modality, we consider an RGB video frame $I \in \mathbb{R}^{H \times W \times 3}$ with spatial dimensions $H \times W$ sampled from a video at time $t$. 
For audio, we use an audio spectrogram $A \in \mathbb{R}^{M \times C}$ spanning several seconds and centered around each video frame at time $t$. %

\textbf{Audio and Image Tokenization.} Following the ViT~\cite{iclr21_vit}, we first decompose each RGB frame $I$  into $n$ non-overlapping patches and then flatten these patches into visual embeddings $\mathbf{X}^{(0)}_v \in \mathbb{R}^{n \times d}$ 
Similarly, we also project audio spectrograms $A$ into audio embeddings $\mathbf{X}^{(0)}_a \in \mathbb{R}^{k \times d}$. 
Note that we inflate the input channel of the audio spectrogram from 1 to 3 to match the dimensions of a linear patch projection layer in the frozen ViT. 

\subsection{Adding \lav Adapters to a Frozen ViT}
\vspace{\subsecmargin}

Next, we describe how we augment a pretrained ViT with our proposed \lav adapters. Every layer of a pretrained ViT in our framework consists of three main operations: (i) a multi-head attention (MHA)~\cite{nips17_attention}, (ii) a multi-layer perceptron (MLP), and (iii) our \lav adapter. As illustrated in~\figref{method}, we add two \lav adapters to every layer in the visual stream and audio stream (i.e., $4$ \lav adapters per layer). Note that every adapter module has its trainable parameters, i.e., the parameters in the adapter modules are not shared. Furthermore, to allow cross-modal exchange, our \lav adapters can transfer information from audio to visual tokens and conversely from visual to audio tokens. Such a bidirectional exchange of information ensures that both modalities aid each other in maximizing the performance of a downstream audio-visual task.

%
\textbf{Standard ViT Layer.} Before describing how to inject \lav adapters into a frozen ViT, we first review how a standard ViT layer processes audio and visual inputs independently. Formally, given audio $\mathbf{X}^{(\ell)}_a$ and visual $\mathbf{X}^{(\ell)}_v$ inputs from a layer $\ell$, the standard ViT layer first independently applies MHA for the inputs from each modality:
\begin{equation}
\begin{aligned}
\label{eq:mha_layer}
\vspace{\eqmargin}
\mathbf{Y}_a^{(\ell)} = \mathbf{X}^{(\ell)}_a + \mathrm{MHA}(\mathbf{X}^{(\ell)}_a),\\
\mathbf{Y}_v^{(\ell)} = \mathbf{X}^{(\ell)}_v + \mathrm{MHA}(\mathbf{X}^{(\ell)}_v).
\end{aligned}
\vspace{\eqmargin}
\end{equation}
%
For brevity, we skip the linear normalization layers in both MHA and MLP operations. Furthermore, for completeness, we define the MHA operation below:
\begin{equation}
\begin{aligned}
\label{eq:mha_eq}
\vspace{\eqmargin}
\text{MHA}(\mathbf{X})=\mathrm{Softmax}\left( (\mathbf{X}\mathbf{W}_{q})(\mathbf{X}\mathbf{W}_{k})^\top\right)(\mathbf{X}\mathbf{W}_{v}).
\end{aligned}
\vspace{\eqmargin}
\end{equation}
Here, $\mathbf{X}$ denotes an input tensor, and $\mathbf{W}_{q},\mathbf{W}_{k},\mathbf{W}_{v} \in \mathbb{R}^{d \times d}$ depict the learnable projection weights.
Afterward, the intermediate representations $\mathbf{Y}_a^{(\ell)}$, and $\mathbf{Y}_v^{(\ell)}$ obtained from the MHA layer are fed into an MLP: 
\begin{equation}
\begin{aligned}
\label{eq:mlp_layer}
\vspace{\eqmargin}
\mathbf{X}_a^{(\ell+1)} = \mathbf{Y}^{(\ell)}_a + \mathrm{MLP}(\mathbf{Y}^{(\ell)}_a),\\
\mathbf{X}_v^{(\ell+1)} = \mathbf{Y}^{(\ell)}_v + \mathrm{MLP}(\mathbf{Y}^{(\ell)}_v).
\end{aligned}
\vspace{\eqmargin}
\end{equation}
The above-defined MHA and MLP operations are then repeatedly applied to audio and visual inputs in each layer of a ViT. With this formal description, we can now describe how to incorporate \lav adapters into a frozen ViT.

\textbf{ViT Layer with a \lav Adapter.} As mentioned above, our model consists of two types of \lav adapters: (i) audio-to-visual (A2V) and (ii) visual-to-audio (V2A). We first describe how to inject an A2V \lav adapter into a frozen ViT. 

Let $\mathbf{F}_v^{(\ell)} = \mathrm{LAV}(\mathbf{X}_a^{(\ell)}, \mathbf{X}_v^{(\ell)})$ denote an operation that implements an audio-to-visual \lav adapter, which we will describe in the next subsection. Then, the updated MHA and MLP operations in each layer can be written as:
\begin{equation}
\begin{aligned}
\label{eq:mha_a2v}
\vspace{\eqmargin}
\mathbf{Y}_v^{(\ell)} = \mathbf{X}^{(\ell)}_v + \mathrm{MHA} (\mathbf{X}^{(\ell)}_v) + \mathrm{LAV}(\mathbf{X}_a^{(\ell)}, \mathbf{X}_v^{(\ell)}),\\
\mathbf{X}_v^{(\ell+1)} = \mathbf{Y}^{(\ell)}_v + \mathrm{MLP}(\mathbf{Y}^{(\ell)}_v) + \mathrm{LAV}(\mathbf{Y}_a^{(\ell)}, \mathbf{Y}_v^{(\ell)}).
\end{aligned}
\vspace{\eqmargin}
\end{equation}
Conceptually, the operation above enables a frozen ViT to incorporate audio features into the visual representation. 

Afterward, we can define a similar formulation for injecting a visual-to-audio (V2A) \lav adapter into a frozen ViT. Let $\mathbf{F}_a^{(\ell)} = \mathrm{LAV}(\mathbf{X}_v^{(\ell)}, \mathbf{X}_a^{(\ell)})$ depict an operation that implements a visual-to-audio \lav adapter, which we will also describe in the next subsection. Then, we can re-write the original MHA and MLP operations (i.e., Equations~\ref{eq:mha_layer},\ref{eq:mlp_layer}) for audio inputs as:
\begin{equation}
\begin{aligned}
\label{eq:mha_v2a}
\vspace{\eqmargin}
\mathbf{Y}_a^{(\ell)} = \mathbf{X}^{(\ell)}_a + \mathrm{MHA} (\mathbf{X}^{(\ell)}_a) + \mathrm{LAV}(\mathbf{X}_v^{(\ell)}, \mathbf{X}_a^{(\ell)}),\\
\mathbf{X}_a^{(\ell+1)} = \mathbf{Y}^{(\ell)}_a + \mathrm{MLP}(\mathbf{Y}^{(\ell)}_a) + \mathrm{LAV}(\mathbf{Y}_v^{(\ell)}, \mathbf{Y}_a^{(\ell)}).
\end{aligned}
\vspace{\eqmargin}
\end{equation}
Intuitively, the operation above allows a frozen ViT to fuse information from the audio and visual tokens for a more expressive audio representation.

\subsection{\lav Adapter}
\vspace{\subsecmargin}
\label{sec:LAVISH}
Lastly, we provide a technical description of our \lav adapter. In a nutshell, \lav adapter is a dual-pathway module that uses a small number of latent tokens to efficiently inject visual cues into the audio representation and vice-versa. It consists of four main technical components: (i) a separate set of latent tokens for audio and visual modalities, (ii) cross-modal attention between audio/visual tokens and latent tokens to compress all tokens of one modality into the latent tokens, (iii) a second cross-modal attention for efficient audio-visual fusion, (iv) a lightweight adapter module that incorporates audio-visual cues into a newly computed feature representation via a small number of trainable parameters. We now describe each of these components in more detail. A detailed illustration of our \lav adapter is presented in~\figref{method}.

\textbf{Latent Tokens.} Inspired by the success of several prior methods~\cite{nips21_bottleneck,perceiver}, we introduce a small set of randomly initialized  latent audio and visual tokens  $\mathbf{L}^{(l)}_a \in \mathbb{R}^{m \times d}$, and $\mathbf{L}^{(l)}_v \in \mathbb{R}^{m \times d}$ respectively. We use a unique set of latent tokens at each layer $l$. Here, $m$ depicts the number of latent tokens, which is significantly smaller than the total number of audio or visual tokens. For instance, the Swin~\cite{cvpr22_swinv2} transformer contains $>2K$ audio or visual tokens. In contrast, in most cases, we use $m=2$ latent tokens, which is orders of magnitude smaller. The purpose of these latent tokens is to compactly summarize information from all the audio and visual tokens for efficient information transfer from one modality to another.

\begin{figure*}[t!]
    \centering
	\includegraphics[width=1\linewidth]{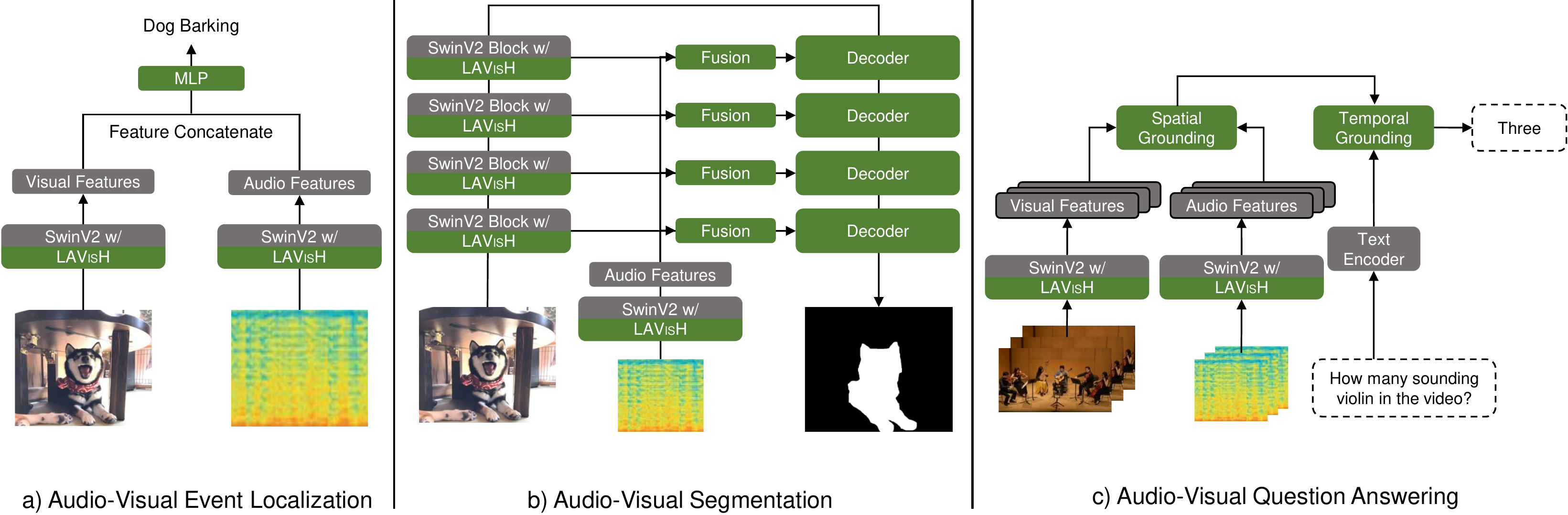}
     \caption{\textbf{Adapting \lav to the Downstream Audio-Visual Tasks} of audio-visual event localization, audio-visual segmentation, and audio-visual question answering.
    The modules \textcolor{cadmiumgreen}{in green} are trainable modules from the baselines~\cite{eccv22_loc_avs,cvpr22_avqa_avqa} that we adopt. Note that the visual and audio backbones in our framework are frozen and share the same parameters. \vspace{-0.4cm}
    }
    \vspace{\figmargin}
	\label{fig:lavish_ada}
\end{figure*}

\textbf{Cross-modal Attention.} We use cross-modal attention (CMA) to implement: (i) a compression module to condense all tokens from one modality into the latent tokens of the same modality and (ii) an audio-visual fusion module, which fuses information between the compressed latent tokens of one modality and all the tokens of the other modality. We define the cross-modal attention operation as:  
\begin{equation}
\begin{aligned}
\label{eq:cm_att}
\vspace{\eqmargin}
\mathrm{CMA}(\mathbf{Q},\mathbf{K},\mathbf{V}) =  \mathbf{Q} + g \cdot \mathrm{Softmax}\left( \mathbf{Q}\mathbf{K}^\top \right) \mathbf{V},
\end{aligned}
\vspace{\eqmargin}
\end{equation}
where $g$ is a learnable scalar to control the flow from one modality to another, and  $\mathbf{Q}$, $\mathbf{K}$, and $\mathbf{V}$ denote query, key, and value tokens respectively. 

\textbf{Audio-Visual Latent Token Compression.} As illustrated in Figure~\ref{fig:method}, we first use cross-modal attention to compress all the visual or audio tokens $\mathbf{X}^{(\ell)}_a$ or $\mathbf{X}^{(\ell)}_v$ into a small set of latent tokens $\mathbf{L}^{(l)}_a$ or $\mathbf{L}^{(l)}_v$ respectively. Formally, this can be written as: 
\begin{equation}
\begin{aligned}
\label{eq:av2l}
\vspace{\eqmargin}
\mathbf{S}^{(\ell)}_{a} & =  \mathrm{CMA}(\mathbf{L}^{(l)}_a,\mathbf{X}^{(\ell)}_a,\mathbf{X}^{(\ell)}_a), \\
\mathbf{S}_v^{(\ell)} &=  \mathrm{CMA}(\mathbf{L}^{(l)}_v,\mathbf{X}^{(\ell)}_v,\mathbf{X}^{(\ell)}_v), \\
\end{aligned}
\vspace{\eqmargin}
\end{equation}
where $\mathbf{S}_a^{(\ell)} \in \mathbb{R}^{m \times d}$ and $\mathbf{S}_v^{(\ell)} \in \mathbb{R}^{m \times d}$ are the latent summary tokens for audio and visual modalities respectively. Intuitively, this operation allows us to compute latent summary tokens $\mathbf{S}_a^{(\ell)}$ and $\mathbf{S}_v^{(\ell)}$ as a weighted summation of all the audio or visual tokens respectively. Furthermore, because the number of latent audio and visual tokens is so small, this forces the model to include only the most relevant audio or visual information into the latent tokens. This in turn enables an efficient cross modal fusion between audio and visual tokens, which we describe next.

\textbf{Audio-Visual Feature Fusion.} We can use the latent summary tokens  $\mathbf{S}_a^{(\ell)}$ and $\mathbf{S}_v^{(\ell)}$ to efficiently fuse information between audio and visual modalities. Formally, we can write this operation as:
\begin{equation}
\begin{aligned}
\label{eq:av2va}
\vspace{\eqmargin}
\mathbf{X}^{(\ell)}_{av} &=  \mathrm{CMA}(\mathbf{X}^{(l)}_a,\mathbf{S}^{(\ell)}_v,\mathbf{S}^{(\ell)}_v), \\
\mathbf{X}^{(\ell)}_{va} &=  \mathrm{CMA}(\mathbf{X}^{(l)}_v,\mathbf{S}^{(\ell)}_a,\mathbf{S}^{(\ell)}_a), \\
\end{aligned}
\vspace{\eqmargin}
\end{equation}
where $\mathbf{X}^{(\ell)}_{av}$ depicts a newly computed audio representation that also incorporates visual cues, and similarly, $\mathbf{X}^{(\ell)}_{va}$ denotes a new visual representation that incorporates audio cues. At a high level, both audio-visual representations  $\mathbf{X}^{(\ell)}_{av}$ and  $\mathbf{X}^{(\ell)}_{va}$ are computed as a weighted combination of the latent summary tokens $\mathbf{S}_v^{(\ell)}$ and $\mathbf{S}_a^{(\ell)}$ respectively. As discussed above, performing cross-modal attention between audio or visual and the latent summary tokens is beneficial because it allows us to avoid the quadratic cost of standard cross-attention operation, which would be very costly due to a large number (i.e., $>2$K) of audio/visual tokens. The resulting audio-visual representations $\mathbf{X}^{(\ell)}_{av}$ and  $\mathbf{X}^{(\ell)}_{va}$ allow both modalities to benefit from each other when solving complex audio-visual understanding tasks.

\textbf{Lightweight Adapter Module.} Following prior work on adapters~\cite{Adapter}, we use a similar bottleneck module that consists of a learnable down-projection layer $\theta_{down}$, a non-linear activation function $\sigma$, and a learnable up-projection layer $\theta_{up}$. The entire  operation can be written as:

\begin{equation}
\begin{aligned}
\label{eq:bot}
\vspace{\eqmargin}
\mathbf{Z}^{(\ell)}_{av} = \theta_{up}(\sigma(\theta_{down}(\mathbf{X}^{(\ell)}_{av}))),\\
\mathbf{Z}^{(\ell)}_{va} = \theta_{up}(\sigma(\theta_{down}(\mathbf{X}^{(\ell)}_{va}))).
\end{aligned}
\vspace{\eqmargin}
\end{equation}

\textbf{Putting It All Together.}  With all the formal definitions above, we can define the final \lav adapter as a sequential application of the three above-described operations: (i) audio-visual latent token compression (Equation~\ref{eq:av2l}), (ii) audio-visual fusion (Equation~\ref{eq:av2va}), and (iii) the lightweight adapter module (Equation~\ref{eq:bot}). Note that these operations are distinct for the visual and audio  inputs. For example, the \lav adapter operation $\mathrm{LAV}(\mathbf{X}_a^{(\ell)}, \mathbf{X}_v^{(\ell)})$ incorporates audio cues into the visual features whereas $\mathrm{LAV}(\mathbf{X}_v^{(\ell)}, \mathbf{X}_a^{(\ell)})$ injects visual cues into the audio features.

\section{Experimental Setup}
\label{sec:exp_setup}

\begin{table*}[t]
    \caption{
    \textbf{Audio-Visual Event Localization.} We compare our proposed \lav approach with previous audio-visual event localization methods.
    \bluexmark~indicates not using an external audio encoder or large-scale audio pretraining.
    In our case, this means that both audio and visual inputs are processed using a visual encoder.
    The~\redfire~and~\cyansnow~denote fully fine-tuned and frozen encoders, respectively.
    $*$ denotes our improved implementations, and $\dagger$~ means that no official code was provided to report some of the baseline-specific metrics.
    The performance is evaluated using audio-visual event classification accuracy.
    Despite not using an external audio encoder or large-scale audio pretraining,  \lav achieves better accuracy than all prior methods while also using a relatively small number of trainable parameters.
    }
    \centering
    \resizebox{0.9\textwidth}{!}{
    \small
    \begin{tabular}{l c c c c c c c}
        \toprule
        Method & \makecell{Visual \\ Encoder} & \makecell{Audio \\ Encoder} & \makecell{Visual \\ Pretrain Dataset}& \makecell{Audio \\ Pretrain Dataset} & \makecell{Trainable \\ Params (M) $\downarrow$} & \makecell{Total \\ Params (M) $\downarrow$} & Acc $\uparrow$\\
       \toprule
       AVT~\cite{my_accv20_av-trans}              & VGG-19 \cyansnow              & VGGish 
 \cyansnow         & ImageNet     & AudioSet           & 15.8     & 231.5       & 76.8 \\
       PSP~\cite{cvpr21_ave_psp}                  & VGG-19 \cyansnow              & VGGish \cyansnow         & ImageNet     & AudioSet           & \bf 1.7     & 217.4       & 77.8 \\
       $\text{DPNet}^\dagger$~\cite{eccv22_ave_DPNet}              & VGG-19            & VGGish      & ImageNet     & AudioSet           & N/A     & N/A       & 79.7 \\
      \midrule
       AVEL~\cite{eccv18_avel}                & ResNet-152 \cyansnow              & VGGish \cyansnow         & ImageNet     & AudioSet            & 3.7     & 136.0       & 74.0 \\
       AVSDN~\cite{my_icassp}                 & ResNet-152 \cyansnow             & VGGish \cyansnow        & ImageNet     & AudioSet            & 8.0  &    140.3    & 75.4 \\
       CMRAN~\cite{acmmm20_ave_CMRAN}         & ResNet-152  \cyansnow             & VGGish  \cyansnow        & ImageNet     & AudioSet            & 15.9     & 148.2       & 78.3 \\
       MM-Pyramid~\cite{acmmm22_ave_MM-Pyramid}         & ResNet-152  \cyansnow             & VGGish  \cyansnow        & ImageNet     & AudioSet            & 44.0     & 176.3      & 77.8 \\
       CMBS~\cite{cvpr22_ave_cmbs}              & ResNet-152 \cyansnow              & VGGish \cyansnow         & ImageNet     & AudioSet            & 14.4        & 216.7   & 79.7 \\
       \midrule
       MBT*~\cite{nips21_bottleneck}          & ViT-B-16  \redfire         & AST \redfire        & ImageNet & AudioSet                 & 172     & 172   & 77.8   \\
       MBT*~\cite{nips21_bottleneck}          & ViT-L-16  \redfire         & AST \redfire        & ImageNet & AudioSet                 & 393      & 393   & OOM   \\
      \bf \lav            &  \multicolumn{2}{c}{ViT-B-16 \cyansnow~(shared) }    & ImageNet & \bluexmark           & 4.7   & \bf 107.2  & 75.3 \\
       \bf \lav            & \multicolumn{2}{c}{ViT-L-16 \cyansnow~(shared) }   & ImageNet & \bluexmark           & 14.5   & 340.1   & 78.1 \\
       \midrule
       CMBS*          & Swin-V2-L  \cyansnow         & VGGish \cyansnow        & ImageNet & AudioSet                 & 14.1      & 315.2   & 80.4   \\
       CMBS*           & Swin-V2-L \redfire         &  VGGish \cyansnow       & ImageNet & AudioSet           & 243.1     & 315.2   &  79.6 \\
       \bf \lav            & \multicolumn{2}{c}{Swin-V2-B \cyansnow~(shared) }    & ImageNet & \bluexmark           & 5.0   & 114.2   & 78.8 \\
       \bf \lav            & \multicolumn{2}{c}{Swin-V2-L \cyansnow~(shared) }             & ImageNet & \bluexmark           & 10.1   & 238.8   & \bf 81.1 \\
       \bottomrule
    \end{tabular}
    }
   \label{tab:sota_ave}
   \vspace{\tabmargin}
\end{table*}

\subsection{Downstream Tasks and Datasets}
\vspace{\subsecmargin}
\textbf{Audio-Visual Event Localization} task focuses on recognizing joint audio and visual events throughout multiple time segments in a video.
We evaluate on the AVE~\cite{eccv18_avel} dataset containing $4,143$ videos, where each video duration is $10$ seconds and contains events spanning $28$ categories.
To adapt our approach to this task, for each time segment, we extract audio and visual features using a frozen visual transformer (e.g., ViT or Swin) augmented with \lav adapters. We then concatenate the audio and visual features and attach a linear layer to obtain the final audio-visual event prediction as shown in \figref{lavish_ada} (a).
Similar to prior approaches~\cite{eccv18_avel,cvpr22_ave_cmbs,cvpr21_ave_psp,eccv22_ave_DPNet}, to assess the performance of our method, we compute the fraction of correctly predicted segments and report it as our evaluation metric.

\textbf{Audio-Visual Segmentation} is a recently introduced task that aims to segment objects given the sound. 
We validate our framework on the AVSBench-S4~\cite{eccv22_loc_avs} dataset, which contains $4,932$ videos with manually annotated pixel-wise annotations of audible objects.
To adapt our framework to this task, we replace the pretrained U-Net visual encoder and the pretrained audio feature extractor of AVS~\cite{eccv22_loc_avs} with our frozen transformer augmented with \lav adapters. We then use it as our audio-visual feature extractor (See \figref{lavish_ada} (b)).
To evaluate our approach, we follow the evaluation protocol of AVSBench-S4, which computes the mean Intersection-over-Union (mIoU) of the predicted segmentation and the ground truth masks.

\textbf{Audio-Visual Question Answering (AVQA)} task requires answering questions based on the associations between objects and sounds.
We conduct our experiments on the MUSIC-AVQA dataset~\cite{cvpr22_avqa_avqa}, which contains $9,288$ videos and $45,867$ question-answer pairs.
To adapt our model to the AVQA task, we replace the pretrained visual encoder and the pretrained audio encoder of the baseline in \cite{cvpr22_avqa_avqa} with our frozen transformer augmented with \lav adapters as presented in \figref{lavish_ada} (c).
Following the original AVQA work~\cite{cvpr22_avqa_avqa}, we evaluate our model using the answer prediction accuracy. %

\begin{table*}[t]
    \caption{
    \textbf{Audio-Visual Segmentation.} We evaluate our \lav approach on the AVSBench-S4~\cite{eccv22_loc_avs} dataset for audio-visual segmentation task using the mean intersection over union (mIoU) metric. Our method achieves comparable performance as the state-of-the-art AVS~\cite{eccv22_loc_avs} approach without relying on an external audio encoder or large-scale audio pretraining. \vspace{-0.2cm}
    }
    \label{tab:sota_av_seg}
    \centering
    \resizebox{0.85\textwidth}{!}{
    \begin{tabular}{l ccccccc}
        \toprule
        Method & \makecell{Visual \\ Encoder} & \makecell{Audio \\ Encoder} &  \makecell{Visual \\ Pretrain Dataset}& \makecell{Audio \\ Pretrain Dataset} & \makecell{Trainable \\ Params (M) $\downarrow$}& \makecell{Total \\ Params (M) $\downarrow$} & mIoU $\uparrow$ \\
        \midrule
        $\text{LVS}^\dagger$~\cite{cvpr21_loc_lvs}      & ResNet18 & ResNet18 & ImageNet& \bluexmark& N/A &N/A& 37.9\\
        $\text{MMSL}^\dagger$~\cite{eccv20_loc_MMSL}    & ResNet-18 & CRNN & ImageNet&AudioSet & N/A&N/A& 44.9\\
        AVS~\cite{eccv22_loc_avs}      & PVT-V2 \redfire & VGGish \cyansnow  & ImageNet&AudioSet & 102.4&\bf 174.5 & 78.7\\
        AVS*    & Swin-V2-L \redfire  & VGGish \cyansnow     & ImageNet&AudioSet & 249.7&321.8 & \bf 80.4\\
        \midrule
         \bf \lav  & \multicolumn{2}{c}{Swin-V2-L \cyansnow~(shared) }  & ImageNet & \bluexmark& \bf 37.2&266.4 & 80.1\\
        \bottomrule
    \end{tabular}
}
\vspace{-2mm}
\end{table*}

\begin{table*}[t]
    \caption{
    \textbf{Audio-Visual Question Answering} on the Music-AVQA~\cite{cvpr22_avqa_avqa} dataset.
    We report accuracy on 3 types of questions, e.g., audio (A), visual (V), and audio-visual (AV).
    Our approach achieves the best accuracy across all three categories of questions including audio-only questions. This verifies the effectiveness of frozen ViT augmented with our \lav adapters to generalize to audio-visual data. \vspace{-0.2cm}  
    }
    \centering
    \resizebox{0.99\textwidth}{!}{
    \small
    \begin{tabular}{l c c c c c c ccc}
        \toprule
        \
        
        \multirow{2}{*}{Method} & \multirow{2}{*}{\makecell{Visual \\ Encoder}}  & \multirow{2}{*}{\makecell{Audio \\ Encoder}} & \multirow{2}{*}{\makecell{Visual \\ Pretrain Dataset}}  & \multirow{2}{*}{\makecell{Audio \\ Pretrain Dataset}} & \multirow{2}{*}{\makecell{Trainable  \\Params (M) $\downarrow$}}& \multirow{2}{*}{\makecell{Total \\ Params (M) $\downarrow$} }& \multicolumn{3}{c}{Question $\uparrow$}\\ 
        \cline {8-10}
        & & & & & & & A & V & AV \\
       \toprule
       $\text{AVSD}^\dagger$~\cite{cvpr19_avqa_avsd}                                     & VGG-19              & VGGish    & ImageNet&AudioSet          & N/A  & N/A  & 68.52 &  70.83   & 65.49 \\
       $\text{Pano-AVQA}^\dagger$~\cite{iccv21_avqa_pano_avqa}                           & Faster RCNN         & VGGish    & ImageNet&AudioSet          & N/A  &N/A  & 70.73 &  72.56  & 66.64 \\
       AVQA~\cite{cvpr22_avqa_avqa}                                     & ResNet-18 \cyansnow  & VGGish \cyansnow         & ImageNet &AudioSet          &\bf 10.6&\bf 94.4   & 74.06 & 74.00   & 69.54 \\
       AVQA*              & Swin-V2-L  \cyansnow            & VGGish \cyansnow         & ImageNet&AudioSet          & 12.23&312.1  & 75.46 & 75.64   & 74.51 \\
       AVQA*              & Swin-V2-L  \redfire           & VGGish \cyansnow         & ImageNet & AudioSet          & 240 & 312.1    & 73.16 & 73.80   & 73.16 \\
       \midrule
       \bf \lav             & \multicolumn{2}{c}{Swin-V2-L \cyansnow~(shared) }    & ImageNet&  \bluexmark           & 21.09 &249.8  & \bf 77.15& \bf 77.37   & \bf 77.08 \\
       \bottomrule
    \end{tabular}
    }
    \label{tab:sota_avqa}
    \vspace{\tabmargin}
    \vspace{-2mm}
\end{table*}


%

\section{Results and Analysis}
\vspace{\secmargin}
\label{sec:results}
\subsection{Audio-Visual Event Localization}
\vspace{\subsecmargin}
In \tabref{sota_ave}, we evaluate our model on the audio-visual event localization task using the AVE~\cite{eccv18_avel} dataset. 
For our main comparisons, we focus on the recent CMBS~\cite{cvpr22_ave_cmbs} method, which achieves state-of-the-art results on this benchmark.
For a fair comparison, we additionally implement this baseline using a Swin-V2-L~\cite{cvpr22_swinv2} backbone, which is also the backbone we use in our \lav approach. 
We also include a modality-specific multimodal fusion bottleneck (MBT) baseline~\cite{nips21_bottleneck} with cross-modal fusion between audio and visual encoders (\ie ViT and AST~\cite{TASLP_AST}) pretrained separately on large-scale image and audio datasets.

Our results in \tabref{sota_ave} indicate several interesting findings. First, we note that, unlike prior approaches~\cite{cvpr22_ave_cmbs,eccv22_ave_DPNet,acmmm20_ave_CMRAN}, our framework does not require a pretrained audio encoder or large-scale audio pretraining on AudioSet~\cite{icassp17_audioset}.
Despite not using a pretrained audio encoder or large-scale AudioSet pretraining, our approach achieves better accuracy ($\bf81.1\%$ vs. $\bf 80.4\%$) than the state-of-the-art CMBS with the Swin-V2-L visual backbone  while also requiring fewer trainable parameters (\textbf{10.1M} vs \textbf{14.1M}).
We also note that the base variant of the modality-specific dual encoder MBT~\cite{nips21_bottleneck} (MBT-B) achieves better performance than \lav with ViT-B encoder (\textbf{77.8\%} vs \textbf{75.3\%}). However, the MBT approach has $\bf{37\times}$ more trainable parameters (\textbf{172M} vs \textbf{4.7M}). Due to the small number of trainable parameters, our approach can be scaled up much more easily than MBT. Specifically, we note that the large MBT variant (MBT-L) requires \textbf{393M} trainable parameters, which leads to the out of memory issues on a 48GB A6000 GPU. In comparison, the large variant of our \lav approach only requires \textbf{14.5M} trainable parameters, which enables memory-efficient training and inference, while also achieving higher accuracy than the best performing MBT variant (\textbf{78.1\%} vs \textbf{77.8\%}).
Lastly, we also observe that Swin-based variants of our model achieve consistently better accuracy than the ViT-based variants (\textbf{81.1\%} vs \textbf{79.6\%}). We hypothesize that since audio information in  spectrograms may be more local than in images, the locality preservation mechanism of Swin may better capture sounds with similar frequencies.

\subsection{Audio-Visual Segmentation}
\vspace{\subsecmargin}

In \tabref{sota_av_seg}, we also evaluate our \lav approach on the audio-visual segmentation task~\cite{eccv22_loc_avs} on the AVSBench-S4~\cite{eccv22_loc_avs} dataset.
Based on our results, we first observe that our framework outperforms the previous best AVS method~\cite{eccv22_loc_avs}  (\textbf{80.1\%} vs \textbf{78.7\%}) while using fewer trainable parameters (\textbf{37.2M} vs. \textbf{249.7M}) and without using an external audio encoder or large-scale audio pretraining.
To make the comparison to the AVS baseline more thorough, we also implement it using the same Swin-V2-L backbone used by our \lav method. In this setting, AVS achieves similar performance to our approach (\textbf{80.4\%} vs. \textbf{80.1\%}). However, this AVS variant uses significantly more trainable parameters than our method (\textbf{249.7M} vs. \textbf{37.2M}). Thus, these results indicate that a frozen transformer augmented with our \lav adapters can generalize to complex dense-prediction tasks such as audio-visual segmentation.

\subsection{Audio-Visual Question Answering}
\vspace{\subsecmargin}

Finally, in \tabref{sota_avqa}, we evaluate our framework on MUSIC-AVQA~\cite{cvpr22_avqa_avqa}, which is an audio-visual question-answering dataset containing three categories of questions (audio, visual, and audio-visual) to assess each method's reasoning capabilities across different modalities.  We compare our \lav approach with the AVSD~\cite{cvpr19_avqa_avsd}, Pano-AVQA~\cite{iccv21_avqa_pano_avqa} and AVQA~\cite{cvpr22_avqa_avqa} methods. Like in the previous tasks, we implement a stronger AVQA baseline using a frozen Swin-V2-L backbone (\ie the same as for our visual encoder). Based on these results, we first observe that our proposed method outperforms all prior approaches by a large margin for all three types of questions (\textbf{+3.09\%}, \textbf{+3.37\%}, and \textbf{+7.54\%}).
Interestingly, we notice that despite not using a pretrained audio encoder or large-scale audio pretraining, \lav achieves better results not only on the visual and audio-visual questions but also on the audio-based questions. This suggests that pretrained ViTs might capture representations that are useful not only for the image but also for the audio data. (i.e., audio images). 
We also note that \lav exhibits larger performance gains on audio-visual questions (\textbf{+2.57\%}) than on visual (\textbf{+1.73\%}) or audio-based (\textbf{+1.69\%}). This suggests that \lav adapters can effectively fuse information across audio and visual modalities for the AVQA task.
\begin{table*}[t]

\footnotesize
    \vspace{-0.3cm}
        \caption{
    \textbf{Audio-visual Action Recognition.} We evaluate our \lav approach on the UCF101~\cite{ucf101} dataset for audio-visual action recognition task. Compared to prior audio-visual approaches, \lav achieves the best action recognition accuracy while using the smallest number of trainable parameters.
    }
    \centering
    \resizebox{0.95\textwidth}{!}{
    \begin{tabular}{l cccccc}
        \toprule
        {Method}  &\makecell{Visual \\ Encoder}&\makecell{Audio \\Encoder} & \makecell{Pretrain \\ Data} & \makecell{Trainable \\ Params (M) $\downarrow$} & \makecell{Samples \\ per Sec. $\uparrow$ } & \makecell{ Acc $\uparrow$ }  \\  
        \midrule
        XDC~\cite{nips20_avcluster}    &R(2+1)D \redfire       & ResNet-18 \redfire  &  Kinetics-400 (A+V) & 45 & -& 86.8 \\
        AVTS~\cite{av_nips18_coop}     &R(2+1)D \redfire       &  ResNet-18\redfire &  Kinetics-400 (A+V) & 45 & -& 86.2 \\
        GDT~\cite{GDT}          &R(2+1)D \redfire         & ResNet-9 \redfire  &  Kinetics-400 (A+V) & 39.2 & -& 89.3 \\
        MBT~\cite{nips21_bottleneck}         &ViT-B \redfire          &  AST-B\redfire &  Kinetics-400 (V) + AudioSet (A)& 172 & 4.42 & 91.8 \\
        \midrule
        LAVISH         &\multicolumn{2}{c}{ViT-B\cyansnow~(shared)}  & Kinetics-400 (V)& \bf 7.4 & \bf 6.36 &  \bf 92.6  \\
        \bottomrule
        
    \end{tabular}
    }
\label{tab:ucf}
\end{table*}
\subsection{Action Recognition on UCF101.}
In~\tabref{ucf}, we also test our model on UCF101~\cite{ucf101} action recognition. We implement \lav using VideoMAE codebase~\cite{nips22_videomae} pretrained on videos only. Compared to XDC, AVTS, and GDT, all of which used large-scale audio-visual pretraining on Kinetics-400, \lav achieves better results (\textbf{92.6}\% vs. \textbf{86.8}\%, \textbf{86.8}\%, and \textbf{89.3}\%) with fewer trainable parameters (\textbf{7.4M} vs. \textbf{45M} and \textbf{39.2M}) and without any audio-visual pretraining. Our method also outperforms MBT, which uses ViT and AST pretrained on Kinetics and AudioSet, respectively. Overall, all the above results reveal that \lav is a plug-and-play module for diverse audio-visual tasks and architectures.

\subsection{Ablation Studies}
\vspace{\subsecmargin}
\label{sec:ablation}

Next, we investigate how different design choices of our model affect the performance on the Audio-Visual Event Localization (AVE)~\cite{eccv18_avel} dataset.

\begin{table}[t]
    \caption{\textbf{\lav Adapter Design.} 
    We investigate different design choices of our \lav adapter on the audio-visual event localization task. Audio-to-visual (A2V) and visual-to-audio (V2A) indicate cross-modal fusion direction. \textsc{AVisH} is a variant of our approach that has the same implementation but does not use latent tokens. Our results indicate that both bidirectional cross-modal fusion and latent tokens are essential for good performance. 
    }
    \label{tab:abs_av-adapter}
    \centering
    \resizebox{0.3\textwidth}{!}{
    \begin{tabular}{l ccc}
        \toprule
        Method   & \makecell{A2V} & \makecell{V2A}   & \makecell{Acc } $\uparrow$ \\
        \toprule
           \multirow{4}{*}{\textsc{AVisH}}   &\redxmark  &\redxmark     & 77.9  \\
                                            & \cadmiumgreencheck &  \redxmark   & 78.7 \\
                                            &  \redxmark& \cadmiumgreencheck    &  76.1 \\
                                             & \cadmiumgreencheck       & \cadmiumgreencheck    & 79.8 \\
        \midrule
        \multirow{4}{*}{\lav}            &  \redxmark &  \redxmark  & 77.9 \\
                                          & \cadmiumgreencheck  &  \redxmark  & 78.8  \\
                                            &  \redxmark & \cadmiumgreencheck    & 78.7   \\
                                          & \cadmiumgreencheck & \cadmiumgreencheck   & \bf 81.1  \\
        \bottomrule
    \end{tabular}
}
\vspace{\tabmargin}
\vspace{-2mm}
\end{table}

%

\textbf{\lav Adapter Design.} In \tabref{abs_av-adapter}, we investigate the usefulness of bidirectional cross-modal fusion and the importance of latent tokens.
To do this, we first introduce an \textsc{AVisH} baseline that has exactly the same design/implementation as \lav but does not use latent tokens in its cross-attention operations. Instead, it directly performs cross-modal fusion on the original audio and visual tokens, which makes it a lot more costly than our \lav scheme.
Furthermore, to study the importance of bidirectional cross-modal fusion, we compare our final bidirectional \lav approach with the unidirectional variants that only use either audio-to-visual (A2V) or visual-to-audio (V2A) cross-modal fusion, and also a baseline that does not use any cross-modal connections. 

To evaluate the performance of each method, we use audio-visual event classification accuracy. Based on the results, in \tabref{abs_av-adapter}, we first note that the bidirectional cross-modal fusion performs better than the baseline without any cross-modal connections for both \textsc{AVisH} (\textbf{+1.9\%}) and \lav (\textbf{+3.2\%}) methods respectively. Additionally, we observe that the bidirectional variants of \textsc{AVisH} and \lav consistently outperform the unidirectional A2V and V2A baselines (\textbf{+1.1\%} and \textbf{+3.7\%} for  \textsc{AVisH} and \textbf{+2.3\%} and \textbf{+2.4\%} for \lav). This verifies that bidirectional cross-modal fusion enables our model better to incorporate audio and visual cues into its representation. 
We also investigate the importance of latent tokens by comparing \lav directly with \textsc{AVisH}. We observe that \lav outperforms \textsc{AVisH} across both unidirectional (\textbf{+0.1\%} and \textbf{+2.6\%}) and bidirectional variants (\textbf{+1.3\%)}. Thus, these results verify the effectiveness of both bidirectional cross-modal fusion and latent tokens.   

\begin{table}[t]
    \caption{\textbf{Comparison with Other Parameter-Efficient Methods.} 
    All parameter-efficient schemes operate on both audio and visual inputs.
    The CMA column depicts whether the cross-modal attention (CMA) is applied for fusing audio-visual information.
    Based on these results, we report that our \lav approach achieves the best performance while also being reasonably efficient in terms of the number of trainable parameters.
    }
    \label{tab:abs_lora}
    \centering
    \resizebox{0.4\textwidth}{!}{
    \begin{tabular}{l ccc}
        \toprule
        Method  & CMA &\makecell{Trainable \\ Params (M) $\downarrow$}  & \makecell{Acc } $\uparrow$ \\
        \toprule
        Prompt Tuning~\cite{emnlp21_prompt}    &\redxmark& 1.2       & 76.0  \\
        \midrule
        Compacter~\cite{Compacter}    &\redxmark & \bf 3.7       & 78.4  \\
        Compacter~\cite{Compacter}    &\cadmiumgreencheck& 3.7       & 78.6  \\
        \midrule
        LoRA~\cite{Lora}         &\redxmark & 17.7         & 79.0  \\
        LoRA~\cite{Lora}         &\cadmiumgreencheck& 17.7         & 79.7  \\
        \midrule
        Adapter~\cite{Adapter}    &\redxmark & 8.9        & 79.1  \\
        Adapter~\cite{Adapter}    &\cadmiumgreencheck& 8.9        & 79.9  \\
        \midrule
        \bf \lav       & \cadmiumgreencheck & 10.1       & \bf 81.1  \\
        \bottomrule
    \end{tabular}
}
\vspace{\tabmargin}
\vspace{-3mm}
\end{table}

\textbf{Computational Cost Analysis.} 
Next, we compare the efficiency of the previously described bidirectional \textsc{AVisH} and \lav methods using the GFLOPs metric. Note that because the backbone encoder is the same for both approaches, we only measure the computational cost of our introduced \lav modules while excluding the cost of the backbone.  We observe that \lav is $\bf 20\times$ times cheaper than \textsc{AVisH} (\textbf{11} vs. \textbf{217} GFLOPs), and \lav saves about $\bf 20\%$ GPU memory for training. This makes sense because, unlike our approach, the \textsc{AVisH} baseline performs cross-attention between every pair of visual and audio tokens. Due to the quadratic cost of cross-attention and a large number of tokens, this operation is very expensive. In contrast, using a small number of latent tokens (e.g., 2) enables efficient audio-visual fusion in our approach.
%

%

%

%

%

\looseness=-1 \textbf{Comparison to Other Parameter-Efficient Schemes.} In \tabref{abs_lora}, we also compare our \lav adapter with other parameter-efficient methods such as Adapter~\cite{Adapter}, Compacter~\cite{Compacter}, and LoRA~\cite{Lora}.
For each of these baselines, we follow the same training pipeline as for our \lav approach except that we replace our \lav adapters with a corresponding parameter-efficient scheme (e.g., Adapter, Compacter or LoRA).
Our results suggest that \lav outperforms LoRA (\textbf{81.1\%} vs. \textbf{79.1\%}) while also using fewer trainable parameters (\textbf{10.1M} vs. \textbf{17.7M}).
Additionally, we note that although Compacter and Adapter use fewer trainable parameters than \lav (\textbf{10.1M} vs. \textbf{8.9M} and \textbf{3.7M}), their accuracy is substantially lower than for our approach (\textbf{81.1\%} vs. \textbf{79.1\%} and \textbf{78.4\%}).
In sum, compared to other parameter-efficient schemes, our \lav adapter provides better accuracy while still being relatively parameter-efficient.

\begin{figure}[t!]
    \centering
	\includegraphics[width=0.85\linewidth]{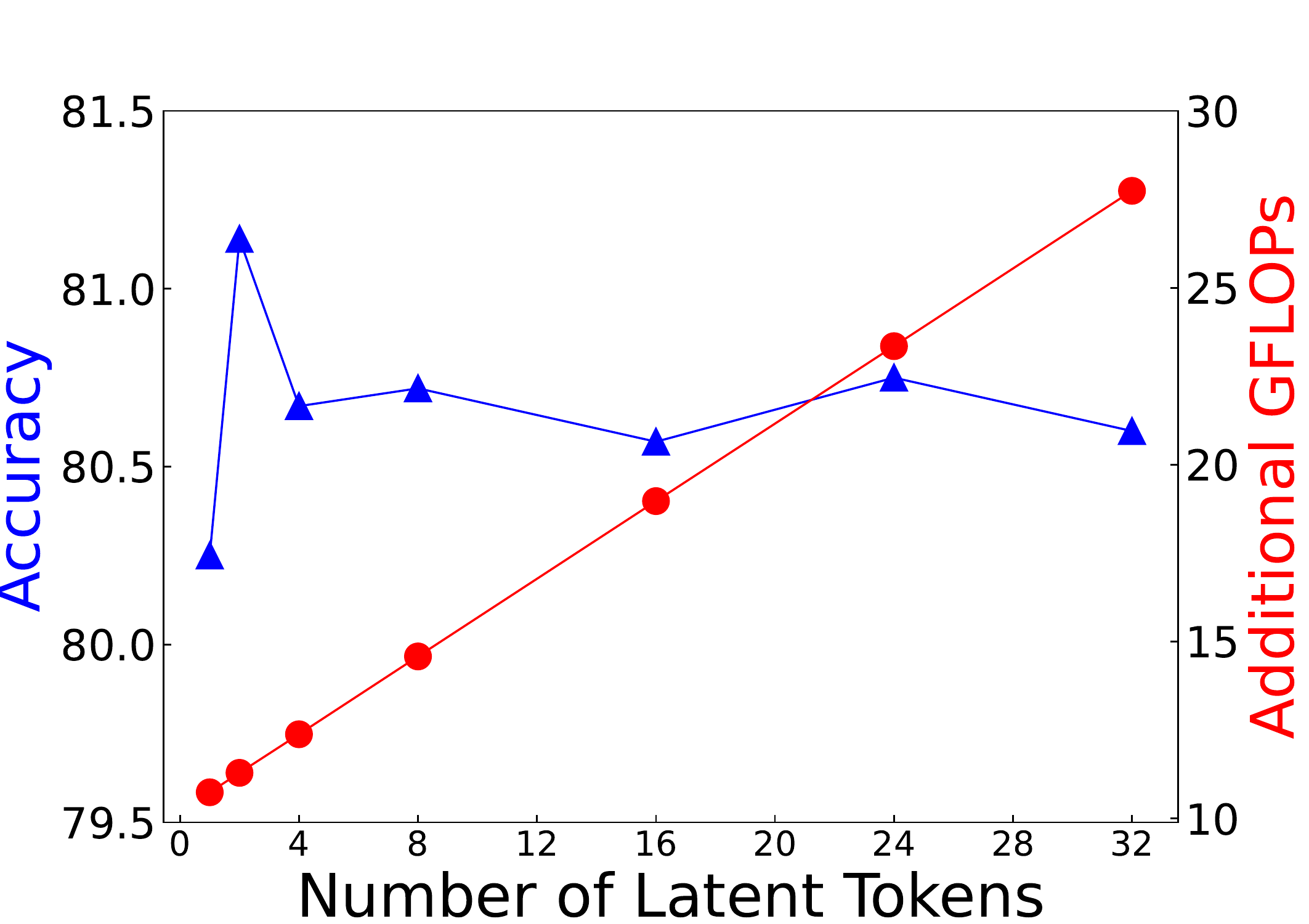}
    \caption{
    \textbf{Number of Latent Tokens.}
    We investigate the accuracy (\textcolor{blue}{in blue}) and the computational cost (in GFLOPs) (\textcolor{red}{in red}) as a function of the number of latent tokens. \lav achieves the best accuracy with two latent tokens. Such a small number of latent tokens enables highly efficient implementation of our approach.\vspace{-0.4cm}
    }
	\label{fig:num_token}
\end{figure}

\begin{table}[t]
    \caption{
    \textbf{Comparison with Visual-only Variants.} We compare our audio-visual approach with visual-only variants on three audio-visual understanding tasks:
    audio-visual event localization (AVE), audio-visual segmentation (AVS), and audio-visual question answering (AVQA). As evaluation metrics, we use top-1 accuracy, mean intersection over union (mIoU), and top-1 accuracy for all three tasks respectively. Our results indicate that our model benefits significantly from jointly modeling audio and visual cues. 
    }
    \centering
    \resizebox{0.4\textwidth}{!}{
    \small
    \begin{tabular}{c c c c }
        \toprule
       \makecell{Task} & \makecell{Input \\ Modality} & \makecell{Accuracy} $\uparrow$&   \\
       \toprule 
       \multirow{2}{*}{AVE~\cite{eccv18_avel}}         & Vision         & 75.3          \\
                 & Audio+Vision            & \bf 81.1             \\
       \midrule
       \multirow{2}{*}{AVS~\cite{eccv22_loc_avs}}   & Vision          & 72.1       \\
           & Audio+Vision            & \bf 80.1           \\
       \midrule
       \multirow{2}{*}{AVQA~\cite{cvpr22_avqa_avqa}}         & Vision          & 63.2        \\
                & Audio+Vision            & \bf 77.1           \\
       \bottomrule
    \end{tabular}
    }
   \label{tab:supp_all_important}

\end{table}

\looseness=-1 \textbf{Number of Latent Tokens.}
Additionally, in \figref{num_token}, we study the performance and computational cost as a function of the number of latent tokens. 
These results indicate that our model achieves the best accuracy with only two tokens (\textbf{81.1\%}).
Furthermore, we observe that using more latent tokens linearly increases the computational cost but does not yield better results. We conjecture that this happens because the AVE dataset is relatively small, and the model might overfit with more latent tokens. This hypothesis is supported by our results on the larger  audio-visual segmentation and audio-visual question answering datasets, where the optimal number of latent tokens is 16. We note that a similar trend has also been reported in prior work~\cite{nips21_bottleneck}.
Thus, these results suggest that \lav obtains a favorable trade-off between performance and efficiency as cross-modal fusion operation can be implemented very efficiently when few (i.e., 2) latent tokens are used.

\looseness=-1 \textbf{Comparison with Visual-Only Baselines.} To verify the importance of jointly considering audio-visual information in all three of our considered benchmarks/tasks (i.e., audio-visual event localization (AVE), audio-visual segmentation (AVS), and audio-visual question-answering (AVQA)), we compare our audio-visual approach with the visual-only variants that only consider visual information without processing any audio cues.
We present these results In \tabref{supp_all_important}, and report the audio-visual variant of our approach, which jointly considers audio and visual cues, consistently outperforms the visual-only baselines by \textbf{5.8\%} top-1 acc., \textbf{8\%} mIoU, and \textbf{13.9\%} top-1 acc. for the AVE, AVS, and AVQA tasks respectively.
These results indicate that our model benefits significantly from the joint modeling of audio and visual cues and also that visual information alone is not enough for achieving state-of-the-art results on these particular audio-visual tasks.

\looseness=-1 \textbf{Comparing ViT and ResNet-152 Backbones.} To investigate whether a visual transformer backbone is truly necessary for adapting a frozen visual model to an audio-visual task, we also conduct experiments with a ResNet-152 backbone~\cite{resnet}. We report that compared to a ViT-B~\cite{iclr21_vit} (86M params), using a ResNet-152 backbone (60M params) leads to a significant \textbf{18}\% drop in accuracy. To make the comparison fairer in terms of a model's capacity, we also report the results using ViT-tiny (6M params) and ViT-small (23M params) architectures, which both have a smaller capacity than ResNet-152. We observe that in both of these cases, the ViT variants outperform ResNet-152 (by \textbf{5.4} \% and \textbf{13.9}\% respectively. These results demonstrate that the lack of inductive biases in visual transformer models enables more effective transfer between inputs across different modalities.

\begin{table*}[t]
    \caption{
    \textbf{Is \lav Complementary to Pretrained Audio Encoders?} 
    We study whether our \lav approach can further benefit from audio features obtained using a VGGish~\cite{icassp17_vggish} audio encoder pretrained on the large-scale AudioSet dataset.
    To do this, we concatenate the pretrained audio features with audio-visual features from our \lav approach. These results indicate that combining audio representations from these two sources leads to a slight boost in performance. %
    }
    \centering
    \resizebox{0.6\textwidth}{!}{
    \small
    \begin{tabular}{l c c c c}
        \toprule
        Method & \makecell{Encoders} & \makecell{Visual \\ Pretrain}& \makecell{Audio \\ Pretrain} & Acc\\
       \toprule
       \lav            & \multicolumn{1}{c}{Swin-V2-L \cyansnow }     & ImageNet & \bluexmark             &  81.1 \\
       \lav             & Swin-V2-L  \cyansnow      ~+~VGGish~\cyansnow    & ImageNet & AudioSet           &  \bf 82.4  \\
       \bottomrule
    \end{tabular}
    }
   \label{tab:supp_comp_vggish}
\end{table*}

\begin{table}[t]
\footnotesize
    \caption{
    \textbf{Throughput Comparison.}
    We compare the throughput of our \lav with the state-of-the-art CMBS approach. The throughput is measured using the number of samples per second. In addition to achieving higher accuracy, our method is almost 2$\times$ faster than CMBS.}
    
    \label{tab:flops}
    \centering
    \resizebox{0.45\textwidth}{!}{
    \begin{tabular}{l cccc}
        \toprule
        {Method}  &\makecell{Visual \\Encoder}&\makecell{Audio \\Encoder} & \makecell{ Samples \\  per Sec. $\uparrow$} & \makecell{ Acc $\uparrow$ }  \\  
        \midrule
        CMBS~\cite{cvpr22_ave_cmbs}     &Swin-L\cyansnow&VGGish\cyansnow                 & 0.72 &  80.4 \\
        LAVISH         &\multicolumn{2}{c}{Swin-L\cyansnow~(shared)}  & \bf 1.40 &    \bf 81.1  \\
        \bottomrule
        
    \end{tabular}
    }
\end{table}

\paragraph{Is \lav Complementary to Pretrained Audio Encoders?}
In~\tabref{supp_comp_vggish}, we also study whether our \lav approach can further benefit from audio features obtained using an external VGGish~\cite{icassp17_vggish} audio encoder pretrained on the large-scale AudioSet dataset. To do this, we concatenate the features from the VGGish~\cite{icassp17_vggish} audio encoder with the audio-visual features from our \lav approach and train a linear layer to predict the event category for the audio-visual event localization task. Based on the results in \tabref{supp_comp_vggish}, we observe that using an external VGGish audio classifier leads to a $1.3\%$ boost in performance. This indicates that our \lav adapters and VGGish encode complementary audio information, and combining audio representations from these two sources is beneficial.

\paragraph{Throughput Comparisons.}
In \tabref{flops}, we also compare LAVISH to CMBS~[79] on the same A6000 GPU. Despite using Swin-L for audio (compared to VGGish), LAVISH has better throughput (\textbf{1.40} vs. \textbf{0.72} samples/sec). This is because, unlike LAVISH, CMBS uses additional temporal modules.

\section{Conclusions}
\vspace{\secmargin}
\label{sec:conclusions}
In this paper, we investigate whether frozen ViTs, pretrained only on images, can generalize to audio-visual data.
We demonstrate that without any audio pretraining our \lav adapter outperforms the state-of-the-art approaches on diverse audio-visual understanding tasks such as audio-visual event localization, audio-visual segmentation, and audio-visual question-answering.
Furthermore, compared to prior approaches, our method requires a significantly smaller number of trainable parameters, enabling efficient audio-visual adaptation.
In the future, we will investigate our model's generalization ability to the audio-only and visual-language tasks, as well as the generalization of pretrained audio models to the audio-visual data.

\section*{Acknowledgments}
We thank Feng Cheng,  Md Mohaiminul Islam, Avinash Madasu, Maitrey Gramopadhye, and Yen-Cheng Liu for helpful discussions. This work was supported by the Sony Faculty Innovation award, ARO Award W911NF2110220, and NSF-AI Engage Institute DRL211263.

\newcount\cvprrulercount
\appendix
\section*{Appendix}

\setcounter{figure}{0}
\setcounter{table}{0}
\renewcommand{\thetable}{A.\arabic{table}}
\renewcommand{\thefigure}{A.\arabic{figure}}
\section{Implementation Details}
For all of our experiments, we extract the visual frames at $1$ fps. 
As our best performing model, we adopt a pretrained Swin-V2-Large~\cite{cvpr22_swinv2} with a $192 \times 192$ spatial resolution with all parameters frozen.
For the audio-visual event localization task, we implement our \lav adapter with $2$ latent tokens and the downsampling factor of $8$ in the 2D group convolutional adapter layers, where the number of group convolutions is set to $2$.
Our group convolution adapter layers use only 0.5x parameters as the standard fully connected ones.
For the audio-visual segmentation and audio-visual question-answering tasks on AVSBench-S4 and MUSIC-AVQA, we use $16$ latent tokens and set the downsampling rate and the number of group convolutions to $4$ and $2$, respectively.
For the audio-visual action recognition on UCF101, we use the same scheme as audio-visual event localization as depicted in \figref{lavish_ada} (a). We use $24$ latent tokens and set the downsampling rate and the number of group convolutions to $4$ and $2$, respectively.
For all of our experiments, we use Adam~\cite{adam} optimizer to train our model.
We set the learning rate of \lav adapter to $5e{-}6$  and $4e{-}6$ for the final prediction layer for audio-visual event localization, $1e{-}4$ for audio-visual segmentation, $8e{-}5$ for \lav adapter and $3e{-}6$ for the grounding modules and the final prediction layer in audio-visual question answering, and $3e{-}5$ for audio-visual action recognition.
For audio preprocessing, we compute the audio spectrogram by PyTorch~\cite{pytorch} kaldi fbank with $192$ triangular mel-frequency bins and frameshift in $5.2$ milliseconds.
Then, we inflate the input channel of the audio spectrogram from 1 to 3 to match the dimensions of a linear patch projection layer in SwinV2.

\begin{table}[h]
\footnotesize
    \vspace{-0.25cm}
    \label{tab:detail}
    \centering
    \resizebox{0.49\textwidth}{!}{
    \begin{tabular}{l ccc}
        \toprule
        {Task}  &\makecell{Batch Size} &\makecell{Num. Latent Tokens}&\makecell{Downsampling  Factor} \\  
        \midrule
        AVE         &2& 2 & 8               \\
        AVS         &4& 16 & 4               \\
        AVQA        &1& 16 & 4               \\
        AVR         &2& 24 & 4               \\
        \bottomrule
        
    \end{tabular}
    }
\vspace{-0.25cm}
\end{table}

{\small
\bibliographystyle{ieee_fullname} 
\bibliography{egbib}

\begin{thebibliography}{100}\itemsep=-1pt

\bibitem{av_eccv20_objs_vids}
Triantafyllos Afouras, Andrew Owens, Joon~Son Chung, and Andrew Zisserman.
\newblock Self-supervised learning of audio-visual objects from video.
\newblock In {\em ECCV}, 2020.

\bibitem{nips21_vatt}
Hassan Akbari, Liangzhe Yuan, Rui Qian, Wei-Hong Chuang, Shih-Fu Chang, Yin
  Cui, and Boqing Gong.
\newblock Vatt: Transformers for multimodal self-supervised learning from raw
  video, audio and text.
\newblock In {\em NeurIPS}, 2021.

\bibitem{alayrac2022flamingo}
Jean-Baptiste Alayrac, Jeff Donahue, Pauline Luc, Antoine Miech, Iain Barr,
  Yana Hasson, Karel Lenc, Arthur Mensch, Katie Millican, Malcolm Reynolds,
  et~al.
\newblock Flamingo: a visual language model for few-shot learning.
\newblock In {\em NeurIPS}, 2022.

\bibitem{nips20_avcluster}
Humam Alwassel, Dhruv Mahajan, Bruno Korbar, Lorenzo Torresani, Bernard Ghanem,
  and Du Tran.
\newblock Self-supervised learning by cross-modal audio-video clustering.
\newblock In {\em NeurIPS}, 2020.

\bibitem{av_eccv18_obj_that_sound}
Relja Arandjelovi{\'c} and Andrew Zisserman.
\newblock Objects that sound.
\newblock In {\em ECCV}, 2018.

\bibitem{iccv21_vivit}
Anurag Arnab, Mostafa Dehghani, Georg Heigold, Chen Sun, Mario Lu{\v{c}}i{\'c},
  and Cordelia Schmid.
\newblock Vivit: A video vision transformer.
\newblock In {\em ICCV}, 2021.

\bibitem{interspeech22_mae_ast}
Alan Baade, Puyuan Peng, and David Harwath.
\newblock Mae-ast: Masked autoencoding audio spectrogram transformer.
\newblock In {\em INTEERSPEECH}, 2022.

\bibitem{iccv21_Frozen}
Max Bain, Arsha Nagrani, G\"ul Varol, and Andrew Zisserman.
\newblock Frozen in time: A joint video and image encoder for end-to-end
  retrieval.
\newblock In {\em ICCV}, 2021.

\bibitem{acl22_bitfit}
Elad Ben~Zaken, Yoav Goldberg, and Shauli Ravfogel.
\newblock {B}it{F}it: Simple parameter-efficient fine-tuning for
  transformer-based masked language-models.
\newblock In {\em ACL}, 2022.

\bibitem{icml21_timesformer}
Gedas Bertasius, Heng Wang, and Lorenzo Torresani.
\newblock Is space-time attention all you need for video understanding?
\newblock In {\em ICML}, 2021.

\bibitem{cvpr21_loc_lvs}
Honglie Chen, Weidi Xie, Triantafyllos Afouras, Arsha Nagrani, Andrea Vedaldi,
  and Andrew Zisserman.
\newblock Localizing visual sounds the hard way.
\newblock In {\em CVPR}, 2021.

\bibitem{icassp20_vggsound}
Honglie Chen, Weidi Xie, Andrea Vedaldi, and Andrew Zisserman.
\newblock Vggsound: A large-scale audio-visual dataset.
\newblock In {\em ICASSP}, 2020.

\bibitem{arxiv22_vit_adapter}
Zhe Chen, Yuchen Duan, Wenhai Wang, Junjun He, Tong Lu, Jifeng Dai, and Yu
  Qiao.
\newblock Vision transformer adapter for dense predictions.
\newblock {\em arXiv Preprint}, 2022.

\bibitem{eccv22_joint_avvp}
Haoyue Cheng, Zhaoyang Liu, Hang Zhou, Chen Qian, Wayne Wu, and Limin Wang.
\newblock Joint-modal label denoising for weakly-supervised audio-visual video
  parsing.
\newblock In {\em ECCV}, 2022.

\bibitem{BERT}
Jacob Devlin, Ming-Wei Chang, Kenton Lee, and Kristina Toutanova.
\newblock Bert: Pre-training of deep bidirectional transformers for language
  understanding.
\newblock In {\em NAACL}, 2018.

\bibitem{iclr21_vit}
Alexey Dosovitskiy, Lucas Beyer, Alexander Kolesnikov, Dirk Weissenborn,
  Xiaohua Zhai, Thomas Unterthiner, Mostafa Dehghani, Matthias Minderer, Georg
  Heigold, Sylvain Gelly, Jakob Uszkoreit, and Neil Houlsby.
\newblock An image is worth 16x16 words: Transformers for image recognition at
  scale.
\newblock In {\em ICLR}, 2021.

\bibitem{iccv19_cosep}
Ruohan Gao and Kristen Grauman.
\newblock Co-separating sounds of visual objects.
\newblock In {\em ICCV}, 2019.

\bibitem{cvpr21_visualvoice}
Ruohan Gao and Kristen Grauman.
\newblock Visualvoice: Audio-visual speech separation with cross-modal
  consistency.
\newblock In {\em CVPR}, 2021.

\bibitem{icassp17_audioset}
Jort~F. Gemmeke, Daniel P.~W. Ellis, Dylan Freedman, Aren Jansen, Wade
  Lawrence, R.~Channing Moore, Manoj Plakal, and Marvin Ritter.
\newblock Audio set: An ontology and human-labeled dataset for audio events.
\newblock In {\em ICASSP}, 2017.

\bibitem{arxiv22_omnimae}
Rohit Girdhar, Alaaeldin El-Nouby, Mannat Singh, Kalyan~Vasudev Alwala, Armand
  Joulin, and Ishan Misra.
\newblock Omnimae: Single model masked pretraining on images and videos.
\newblock {\em arXiv Preprint}, 2022.

\bibitem{cvpr22_omnivore}
Rohit Girdhar, Mannat Singh, Nikhila Ravi, Laurens van~der Maaten, Armand
  Joulin, and Ishan Misra.
\newblock Omnivore: A single model for many visual modalities.
\newblock In {\em CVPR}, 2022.

\bibitem{interspeech21_AST}
Yuan Gong, Yu-An Chung, and James Glass.
\newblock {AST: Audio Spectrogram Transformer}.
\newblock In {\em INTEERSPEECH}, 2021.

\bibitem{TASLP_AST}
Yuan Gong, Yu-An Chung, and James Glass.
\newblock Psla: Improving audio tagging with pretraining, sampling, labeling,
  and aggregation.
\newblock {\em TASLP}, 2021.

\bibitem{arxiv22_uavm}
Yuan Gong, Alexander~H Liu, Andrew Rouditchenko, and James Glass.
\newblock Uavm: A unified model for audio-visual learning.
\newblock {\em arXiv Preprint}, 2022.

\bibitem{acl21_parameter_diff_pruning}
Demi Guo, Alexander~M Rush, and Yoon Kim.
\newblock Parameter-efficient transfer learning with diff pruning.
\newblock In {\em ACL}, 2021.

\bibitem{resnet}
Kaiming He, Xiangyu Zhang, Shaoqing Ren, and Jian Sun.
\newblock Deep residual learning for image recognition.
\newblock In {\em CVPR}, 2016.

\bibitem{icassp17_vggish}
Shawn Hershey, Sourish Chaudhuri, Daniel P.~W. Ellis, Jort~F. Gemmeke, Aren
  Jansen, Channing Moore, Manoj Plakal, Devin Platt, Rif~A. Saurous, Bryan
  Seybold, Malcolm Slaney, Ron Weiss, and Kevin Wilson.
\newblock Cnn architectures for large-scale audio classification.
\newblock In {\em ICASSP}, 2017.

\bibitem{Adapter}
Neil Houlsby, Andrei Giurgiu, Stanislaw Jastrzebski, Bruna Morrone, Quentin
  De~Laroussilhe, Andrea Gesmundo, Mona Attariyan, and Sylvain Gelly.
\newblock Parameter-efficient transfer learning for nlp.
\newblock In {\em ICML}, 2019.

\bibitem{av_nips20_loc}
Di Hu, Rui Qian, Minyue Jiang, Xiao Tan, Shilei Wen, Errui Ding, Weiyao Lin,
  and Dejing Dou.
\newblock Discriminative sounding objects localization via self-supervised
  audiovisual matching.
\newblock In {\em NeurIPS}, 2020.

\bibitem{Lora}
Edward~J Hu, Yelong Shen, Phillip Wallis, Zeyuan Allen-Zhu, Yuanzhi Li, Shean
  Wang, Lu Wang, and Weizhu Chen.
\newblock Lora: Low-rank adaptation of large language models.
\newblock In {\em ICLR}, 2022.

\bibitem{cvpr22_loc_mix}
Xixi Hu, Ziyang Chen, and Andrew Owens.
\newblock Mix and localize: Localizing sound sources in mixtures.
\newblock In {\em CVPR}, 2022.

\bibitem{perceiver}
Andrew Jaegle, Felix Gimeno, Andy Brock, Oriol Vinyals, Andrew Zisserman, and
  Joao Carreira.
\newblock Perceiver: General perception with iterative attention.
\newblock In {\em ICML}, 2021.

\bibitem{Compacter}
Rabeeh Karimi~Mahabadi, James Henderson, and Sebastian Ruder.
\newblock Compacter: Efficient low-rank hypercomplex adapter layers.
\newblock In {\em NeurIPS}, 2021.

\bibitem{adam}
Diederik~P Kingma and Jimmy Ba.
\newblock Adam: A method for stochastic optimization.
\newblock In {\em ICLR}, 2015.

\bibitem{av_nips18_coop}
Bruno Korbar, Du Tran, and Lorenzo Torresani.
\newblock Cooperative learning of audio and video models from self-supervised
  synchronization.
\newblock In {\em NeurIPS}, 2018.

\bibitem{av_iclr21_lee2021crossattentional}
Jun-Tae Lee, Mihir Jain, Hyoungwoo Park, and Sungrack Yun.
\newblock Cross-attentional audio-visual fusion for weakly-supervised action
  localization.
\newblock In {\em ICLR}, 2021.

\bibitem{iclr21_parameter_av}
Sangho Lee, Youngjae Yu, Gunhee Kim, Thomas Breuel, Jan Kautz, and Yale Song.
\newblock Parameter efficient multimodal transformers for video representation
  learning.
\newblock In {\em ICLR}, 2021.

\bibitem{emnlp21_power}
Brian Lester, Rami Al-Rfou, and Noah Constant.
\newblock The power of scale for parameter-efficient prompt tuning.
\newblock In {\em EMNLP}, 2021.

\bibitem{emnlp21_prompt}
Brian Lester, Rami Al-Rfou, and Noah Constant.
\newblock The power of scale for parameter-efficient prompt tuning.
\newblock In {\em EMNLP}, 2021.

\bibitem{cvpr22_avqa_avqa}
Guangyao Li, Yake Wei, Yapeng Tian, Chenliang Xu, Ji-Rong Wen, and Di Hu.
\newblock Learning to answer questions in dynamic audio-visual scenarios.
\newblock In {\em CVPR}, 2022.

\bibitem{acl21_prefix_tuning}
Xiang~Lisa Li and Percy Liang.
\newblock Prefix-tuning: Optimizing continuous prompts for generation.
\newblock In {\em ACL}, 2021.

\bibitem{arxiv21_polyvit}
Valerii Likhosherstov, Anurag Arnab, Krzysztof Choromanski, Mario Lucic, Yi
  Tay, Adrian Weller, and Mostafa Dehghani.
\newblock Polyvit: Co-training vision transformers on images, videos and audio.
\newblock {\em arXiv Preprint}, 2021.

\bibitem{cvpr22_swinbert}
Kevin Lin, Linjie Li, Chung-Ching Lin, Faisal Ahmed, Zhe Gan, Zicheng Liu,
  Yumao Lu, and Lijuan Wang.
\newblock Swinbert: End-to-end transformers with sparse attention for video
  captioning.
\newblock In {\em CVPR}, 2022.

\bibitem{eccv22_eclipse}
Yan-Bo Lin, Jie Lei, Mohit Bansal, and Gedas Bertasius.
\newblock Eclipse: Efficient long-range video retrieval using sight and sound.
\newblock In {\em ECCV}, 2022.

\bibitem{my_icassp}
Yan-Bo Lin, Yu-Jhe Li, and Yu-Chiang~Frank Wang.
\newblock Dual-modality seq2seq network for audio-visual event localization.
\newblock In {\em ICASSP}, 2019.

\bibitem{my_nips21}
Yan-Bo Lin, Hung-Yu Tseng, Hsin-Ying Lee, Yen-Yu Lin, and Ming-Hsuan Yang.
\newblock Exploring cross-video and cross-modality signals for
  weakly-supervised audio-visual video parsing.
\newblock In {\em NeurIPS}, 2021.

\bibitem{my_accv20_av-trans}
Yan-Bo Lin and Yu-Chiang~Frank Wang.
\newblock Audiovisual transformer with instance attention for audio-visual
  event localization.
\newblock In {\em ACCV}, 2020.

\bibitem{eccv22_frozen_clip}
Ziyi Lin, Shijie Geng, Renrui Zhang, Peng Gao, Gerard de Melo, Xiaogang Wang,
  Jifeng Dai, Yu Qiao, and Hongsheng Li.
\newblock Frozen clip models are efficient video learners.
\newblock In {\em ECCV}, 2022.

\bibitem{nips22_few_pe}
Haokun Liu, Derek Tam, Mohammed Muqeeth, Jay Mohta, Tenghao Huang, Mohit
  Bansal, and Colin Raffel.
\newblock Few-shot parameter-efficient fine-tuning is better and cheaper than
  in-context learning.
\newblock In {\em NeurIPS}, 2022.

\bibitem{nips22_polyhistor}
Yen-Cheng Liu, Chih-Yao Ma, Junjiao Tian, Zijian He, and Zsolt Kira.
\newblock Polyhistor: Parameter-efficient multi-task adaptation for dense
  vision tasks.
\newblock In {\em NeurIPS}, 2022.

\bibitem{cvpr22_swinv2}
Ze Liu, Han Hu, Yutong Lin, Zhuliang Yao, Zhenda Xie, Yixuan Wei, Jia Ning, Yue
  Cao, Zheng Zhang, Li Dong, et~al.
\newblock Swin transformer v2: Scaling up capacity and resolution.
\newblock In {\em CVPR}, 2022.

\bibitem{arxiv21_universal_engines}
Kevin Lu, Aditya Grover, Pieter Abbeel, and Igor Mordatch.
\newblock Pretrained transformers as universal computation engines.
\newblock {\em arXiv Preprint}, 2021.

\bibitem{arxiv_clip4clip}
Huaishao Luo, Lei Ji, Ming Zhong, Yang Chen, Wen Lei, Nan Duan, and Tianrui Li.
\newblock {CLIP4Clip}: An empirical study of clip for end to end video clip
  retrieval.
\newblock {\em arXiv Preprint}, 2021.

\bibitem{av_iclr21_activeContrastive}
Shuang Ma, Zhaoyang Zeng, Daniel McDuff, and Yale Song.
\newblock Active contrastive learning of audio-visual video representations.
\newblock In {\em ICLR}, 2021.

\bibitem{nips21_av_contrastive}
Shuang Ma, Zhaoyang Zeng, Daniel McDuff, and Yale Song.
\newblock Contrastive learning of global and local audio-visual
  representations.
\newblock In {\em NeurIPS}, 2021.

\bibitem{acl21_hyperformer}
Rabeeh~Karimi Mahabadi, Sebastian Ruder, Mostafa Dehghani, and James Henderson.
\newblock Parameter-efficient multi-task fine-tuning for transformers via
  shared hypernetworks.
\newblock In {\em ACL}, 2021.

\bibitem{wacv23_AVE_CLIP}
Tanvir Mahmud and Diana Marculescu.
\newblock Ave-clip: Audioclip-based multi-window temporal transformer for audio
  visual event localization.
\newblock In {\em WACV}, 2023.

\bibitem{nips22_slavc}
Shentong Mo and Pedro Morgado.
\newblock A closer look at weakly-supervised audio-visual source localization.
\newblock In {\em NeurIPS}, 2022.

\bibitem{eccv22_loc_ezvsl}
Shentong Mo and Pedro Morgado.
\newblock Localizing visual sounds the easy way.
\newblock In {\em ECCV}, 2022.

\bibitem{nips22_group_avvp}
Shentong Mo and Yapeng Tian.
\newblock Multi-modal grouping network for weakly-supervised audio-visual video
  parsing.
\newblock In {\em NeurIPS}, 2022.

\bibitem{eccv22_nagrani2022learning}
Arsha Nagrani, Paul~Hongsuck Seo, Bryan Seybold, Anja Hauth, Santiago Manen,
  Chen Sun, and Cordelia Schmid.
\newblock Learning audio-video modalities from image captions.
\newblock In {\em ECCV}, 2022.

\bibitem{nips21_bottleneck}
Arsha Nagrani, Shan Yang, Anurag Arnab, Aren Jansen, Cordelia Schmid, and Chen
  Sun.
\newblock Attention bottlenecks for multimodal fusion.
\newblock In {\em NeurIPS}, 2021.

\bibitem{av_eccv18_Owens}
Andrew Owens and Alexei~A. Efros.
\newblock Audio-visual scene analysis with self-supervised multisensory
  features.
\newblock In {\em ECCV}, 2018.

\bibitem{nips22_ST_Adapter}
Junting Pan, Ziyi Lin, Xiatian Zhu, Jing Shao, and Hongsheng Li.
\newblock St-adapter: Parameter-efficient image-to-video transfer learning for
  action recognition.
\newblock In {\em NeurIPS}, 2022.

\bibitem{pytorch}
Adam Paszke, Sam Gross, Francisco Massa, Adam Lerer, James Bradbury, Gregory
  Chanan, Trevor Killeen, Zeming Lin, Natalia Gimelshein, Luca Antiga, et~al.
\newblock Pytorch: An imperative style, high-performance deep learning library.
\newblock In {\em NeurIPS}, 2019.

\bibitem{GDT}
Mandela Patrick, Yuki~M Asano, Polina Kuznetsova, Ruth Fong, Joao~F Henriques,
  Geoffrey Zweig, and Andrea Vedaldi.
\newblock On compositions of transformations in contrastive self-supervised
  learning.
\newblock In {\em ICCV}, 2021.

\bibitem{eccv20_loc_MMSL}
Rui Qian, Di Hu, Heinrich Dinkel, Mengyue Wu, Ning Xu, and Weiyao Lin.
\newblock Multiple sound sources localization from coarse to fine.
\newblock In {\em ECCV}, 2020.

\bibitem{icml21_clip}
Alec Radford, Jong~Wook Kim, Chris Hallacy, Aditya Ramesh, Gabriel Goh,
  Sandhini Agarwal, Girish Sastry, Amanda Askell, Pamela Mishkin, Jack Clark,
  et~al.
\newblock Learning transferable visual models from natural language
  supervision.
\newblock In {\em ICML}, 2021.

\bibitem{icassp20_ave_avin}
Janani Ramaswamy.
\newblock What makes the sound?: A dual-modality interacting network for
  audio-visual event localization.
\newblock In {\em ICASSP}, 2020.

\bibitem{wacv20_ave_avrb}
Janani Ramaswamy and Sukhendu Das.
\newblock See the sound, hear the pixels.
\newblock In {\em WACV}, 2020.

\bibitem{eccv22_ave_DPNet}
Varshanth Rao, Md~Ibrahim Khalil, Haoda Li, Peng Dai, and Juwei Lu.
\newblock Dual perspective network for audio-visual event localization.
\newblock In {\em ECCV}, 2022.

\bibitem{nips17_residual_adapters}
Sylvestre-Alvise Rebuffi, Hakan Bilen, and Andrea Vedaldi.
\newblock Learning multiple visual domains with residual adapters.
\newblock In {\em NeurIPS}, 2017.

\bibitem{cvpr18_efficient_multi-domain}
Sylvestre-Alvise Rebuffi, Hakan Bilen, and Andrea Vedaldi.
\newblock Efficient parametrization of multi-domain deep neural networks.
\newblock In {\em CVPR}, 2018.

\bibitem{cvpr19_avqa_avsd}
Idan Schwartz, Alexander~G Schwing, and Tamir Hazan.
\newblock A simple baseline for audio-visual scene-aware dialog.
\newblock In {\em CVPR}, 2019.

\bibitem{av_cvpr18_lls}
Arda Senocak, Tae-Hyun Oh, Junsik Kim, Ming-Hsuan Yang, and In~So Kweon.
\newblock Learning to localize sound source in visual scenes.
\newblock In {\em CVPR}, 2018.

\bibitem{av_tpami20_lls}
Arda Senocak, Tae-Hyun Oh, Junsik Kim, Ming-Hsuan Yang, and {In}-So Kweon.
\newblock Learning to localize sound sources in visual scenes: Analysis and
  applications.
\newblock {\em TPAMI}, 2019.

\bibitem{iclr22_avhubert}
Bowen Shi, Wei-Ning Hsu, Kushal Lakhotia, and Abdelrahman Mohamed.
\newblock Learning audio-visual speech representation by masked multimodal
  cluster prediction.
\newblock In {\em ICLR}, 2022.

\bibitem{ucf101}
Khurram Soomro, Amir~Roshan Zamir, and Mubarak Shah.
\newblock Ucf101: A dataset of 101 human actions classes from videos in the
  wild.
\newblock {\em arXiv Preprint}, 2012.

\bibitem{nips22_lst}
Yi-Lin Sung, Jaemin Cho, and Mohit Bansal.
\newblock Lst: Ladder side-tuning for parameter and memory efficient transfer
  learning.
\newblock In {\em NeurIPS}, 2022.

\bibitem{cvpr22_vl_adapter}
Yi-Lin Sung, Jaemin Cho, and Mohit Bansal.
\newblock Vl-adapter: Parameter-efficient transfer learning for
  vision-and-language tasks.
\newblock In {\em CVPR}, 2022.

\bibitem{nips21_training_sparse_masks}
Yi-Lin Sung, Varun Nair, and Colin~A Raffel.
\newblock Training neural networks with fixed sparse masks.
\newblock In {\em NeurIPS}, 2021.

\bibitem{nips22_tvlt_textless}
Zineng Tang, Jaemin Cho, Yixin Nie, and Mohit Bansal.
\newblock Tvlt: Textless vision-language transformer.
\newblock In {\em NeurIPS}, 2022.

\bibitem{cvpr21_cyclic}
Yapeng Tian, Di Hu, and Chenliang Xu.
\newblock Cyclic co-learning of sounding object visual grounding and sound
  separation.
\newblock In {\em CVPR}, 2021.

\bibitem{av_eccv20_avvp}
Yapeng Tian, Dingzeyu Li, and Chenliang Xu.
\newblock Unified multisensory perception: Weakly-supervised audio-visual video
  parsing.
\newblock In {\em ECCV}, 2020.

\bibitem{eccv18_avel}
Yapeng Tian, Jing Shi, Bochen Li, Zhiyao Duan, and Chenliang Xu.
\newblock Audio-visual event localization in unconstrained videos.
\newblock In {\em ECCV}, 2018.

\bibitem{nips22_videomae}
Zhan Tong, Yibing Song, Jue Wang, and Limin Wang.
\newblock Video{MAE}: Masked autoencoders are data-efficient learners for
  self-supervised video pre-training.
\newblock In {\em NeurIPS}, 2022.

\bibitem{nips17_attention}
Ashish Vaswani, Noam Shazeer, Niki Parmar, Jakob Uszkoreit, Llion Jones,
  Aidan~N Gomez, {\L}ukasz Kaiser, and Illia Polosukhin.
\newblock Attention is all you need.
\newblock In {\em NeurIPS}, 2017.

\bibitem{av_cvpr21_av_parsing}
Yu Wu and Yi Yang.
\newblock Exploring heterogeneous clues for weakly-supervised audio-visual
  video parsing.
\newblock In {\em CVPR}, 2021.

\bibitem{iccv19_ave_DAM}
Yu Wu, Linchao Zhu, Yan Yan, and Yi Yang.
\newblock Dual attention matching for audio-visual event localization.
\newblock In {\em ICCV}, 2019.

\bibitem{cvpr22_ave_cmbs}
Yan Xia and Zhou Zhao.
\newblock Cross-modal background suppression for audio-visual event
  localization.
\newblock In {\em CVPR}, 2022.

\bibitem{av_slowfast}
Fanyi Xiao, Yong~Jae Lee, Kristen Grauman, Jitendra Malik, and Christoph
  Feichtenhofer.
\newblock Audiovisual slowfast networks for video recognition.
\newblock {\em arXiv Preprint}, 2020.

\bibitem{nips22_audioMAE}
Hu Xu, Juncheng Li, Alexei Baevski, Michael Auli, Wojciech Galuba, Florian
  Metze, Christoph Feichtenhofer, et~al.
\newblock Masked autoencoders that listen.
\newblock In {\em NeurIPS}, 2022.

\bibitem{acmmm20_ave_CMRAN}
Haoming Xu, Runhao Zeng, Qingyao Wu, Mingkui Tan, and Chuang Gan.
\newblock Cross-modal relation-aware networks for audio-visual event
  localization.
\newblock In {\em ACM MM}, 2020.

\bibitem{aaai20_ave_cman}
Hanyu Xuan, Zhenyu Zhang, Shuo Chen, Jian Yang, and Yan Yan.
\newblock Cross-modal attention network for temporal inconsistent audio-visual
  event localization.
\newblock In {\em AAAI}, 2020.

\bibitem{acmmm22_ave_MM-Pyramid}
Jiashuo Yu, Ying Cheng, Rui-Wei Zhao, Rui Feng, and Yuejie Zhang.
\newblock Mm-pyramid: multimodal pyramid attentional network for audio-visual
  event localization and video parsing.
\newblock In {\em ACM MM}, 2022.

\bibitem{iccv21_avqa_pano_avqa}
Heeseung Yun, Youngjae Yu, Wonsuk Yang, Kangil Lee, and Gunhee Kim.
\newblock Pano-avqa: Grounded audio-visual question answering on 360deg videos.
\newblock In {\em ICCV}, 2021.

\bibitem{eccv22_tip_adapter}
Renrui Zhang, Rongyao Fang, Peng Gao, Wei Zhang, Kunchang Li, Jifeng Dai, Yu
  Qiao, and Hongsheng Li.
\newblock Tip-adapter: Training-free clip-adapter for better vision-language
  modeling.
\newblock In {\em arXiv Preprint}, 2022.

\bibitem{cvpr22_ready_audio_adaptive}
Yunhua Zhang, Hazel Doughty, Ling Shao, and Cees~GM Snoek.
\newblock Audio-adaptive activity recognition across video domains.
\newblock In {\em CVPR}, 2022.

\bibitem{eccv22_loc_avs}
Jinxing Zhou, Jianyuan Wang, Jiayi Zhang, Weixuan Sun, Jing Zhang, Stan
  Birchfield, Dan Guo, Lingpeng Kong, Meng Wang, and Yiran Zhong.
\newblock Audio--visual segmentation.
\newblock In {\em ECCV}, 2022.

\bibitem{cvpr21_ave_psp}
Jinxing Zhou, Liang Zheng, Yiran Zhong, Shijie Hao, and Meng Wang.
\newblock Positive sample propagation along the audio-visual event line.
\newblock In {\em CVPR}, 2021.

\end{thebibliography}
}

\end{document}